ARTICLE

# Dataset for Identification of Homophobia and Transophobia in Multilingual YouTube Comments

Bharathi Raja Chakravarthi[•], Ruba Priyadharshini[\], Rahul Ponnusamy[tft], Prasanna Kumar Kumaresan[tft], Kayalvizhi Sampath[♦], Durairaj Thenmozhi [♦], Sathiyaraj Thangasamy[tl], Rajendran Nallathambi[1r], John Phillip McCrae[•]

[•] Insight SFI Research Centre for Data Analytics, National University of Ireland Galway, Galway, Ireland
[\]ULTRA Arts and Science College, Madurai, Tamil Nadu, India
[tft]Indian Institute of Information Technology and Management-Kerala, Thiruvananthapuram, Kerala, India
[♦] Sri Sivasubramaniya Nadar College of Engineering, Chennai, Tamil Nadu, India
[tl]Sri Krishna Adithya College of Arts and Science, Coimbatore, Tamil Nadu, India
[1r]Hindusthan College of Arts and Science, Coimbatore, Tamil Nadu, India
bharathi.raja@insight-centre.org, rubapriyadharshini.a@gmail.com, rahul.mi20@iiitmk.ac.in, prasanna.mi20@iiitmk.ac.in, kayalvizhis@ssn.edu.in, theni_d@ssn.edu.in, sathiyarajt@skacas.ac.in, rajendran.n@hicas.ac.in, john.mccrae@insight-centre.org

**Abstract**

The increased proliferation of abusive content on social media platforms has a negative impact on online users. The dread, dislike, discomfort, or mistrust of lesbian, gay, transgender or bisexual persons is defined as homophobia/transphobia. Homophobic/transphobic speech is a type of offensive language that may be summarized as hate speech directed toward LGBT+ people, and it has been a growing concern in recent years. Online homophobia/transphobia is a severe societal problem that can make online platforms poisonous and unwelcome to LGBT+ people while also attempting to eliminate equality, diversity, and inclusion. We provide a new hierarchical taxonomy for online homophobia and transphobia, as well as an expert-labelled dataset that will allow homophobic/transphobic content to be automatically identified. We educated annotators and supplied them with comprehensive annotation rules because this is a sensitive issue, and we previously discovered that untrained crowdsourcing annotators struggle with diagnosing homophobia due to cultural and other prejudices. The dataset comprises 15,141 annotated multilingual comments. This paper describes the process of building the dataset, qualitative analysis of data, and inter-annotator agreement. In addition, we create baseline models for the dataset. To the best of our knowledge, our dataset is the first such dataset created.

**Warning:** This paper contains explicit statements of homophobia, transphobia, stereotypes which may be distressing to some readers.

## 1. Introduction

Social media plays an essential role in online communication in the digital era, allowing users to post and share material and express their views and thoughts on anything at any time (Gkotsis et al. 2016; Wang et al. 2019). This quantity of data allows researchers in the domains of natural language processing (NLP) to tackle more in-depth long-standing challenges like interpreting, quantifying, and monitoring user behaviour toward certain subjects or events (Engonopoulos et al. 2013). Furthermore, the fast advancement of deep learning-based NLP and the vast amount of user-generated material available online, particularly on social media, enable robust and successful ways to analyse users' behaviours (Yamada et al. 2019). Such strategies can be used for reasons such as gathering affective behaviour analysis or misogyny detection (Tortoreto et al. 2019).

On the Internet, there is a wide range of unpleasant information, including racist, homophobic, transphobic, and sexist messages, as well as insults and threats directed at people or organisations (Zampieri et al. 2019a). As a result of the increase in online material, it has become a significant problem for online communities (Kumar et al. 2018). Online offensive language has been identified as a worldwide phenomenon diffused throughout social media platforms such as Facebook, YouTube, and Twitter during the last decade (Gao et al. 2020). It is even more distressing for Lesbian, Gay, Bisexual, Transgender and other (LGBT+) vulnerable individuals (Díaz-Torres et al. 2020). Because of who they love, how they appear, or who they are, LGBT+ people all across the globe are subjected to violence and inequity, as well as torture and even execution (Barrientos et al. 2010; Schneider and Dimito 2010). Sexual orientation and gender identity are important components of our identities that should never be discriminated against or abused (Thurlow 2001). However, in many countries, being identified as LGBT+ will cost lives, so the vulnerable individual go to social media to get support or share their stories in the hope of finding similar people (Adkins et al. 2018; Han et al. 2019). In recent years, we have seen an explosion in the use of social media, where fundamental standards of conduct constrain freedom of speech in order to maintain a healthy atmosphere and avoid online abuse. However, by utilising the unique qualities of Internet such as anonymity, the Internet has allowed everyone to obtain significant influence over the lives of other individual people. Unfortunately, LGBT+ people who find solace on the Internet are also targeted by homophobic or transphobic attacks. As a result, LGBT+ people who come online for support are being attacked or abused, causing severe mental health issues (McConnell et al. 2017; Wright and Wachs 2021).

Automatic identification of homophobic and transphobic language on the Internet might make it easier to ban anti-LGBT+ harmful materials and move the Internet towards equality, diversity, and inclusion. While much work has been put into the domain of aggression identification (Risch and Krestel 2018), misogyny (Fersini et al. 2020; Bhattacharya et al. 2020), and racism (Waseem 2016), homophobic or transphobic verbal abuse, on the other hand, was given as far less important than racist or other prohibited issues. The absence of homophobic and transphobic annotated data has hampered the development of homophobic and transphobic detection algorithms. Vulnerable LGBT+ people in India, like their counterparts throughout the world, are subjected to many forms of online abuse, which reinforces and justifies homophobic attack, inferior social standing, sexual assault, assault, maltreatment, and disrespect (Rao and Jacob 2014; Chakrapani et al. 2017; Kealy-Bateman 2018; Kar 2018; Billies et al. 2009; Chauhan et al. 2021). Furthermore, in some situations, online abuse toward these vulnerable individuals may grow into systematic bullying campaigns initiated on social media to target and, in some cases, brutally threaten LGBT+ who are popular in social media (Garaigordobil et al. 2020; Mkhize et al. 2020; Ybarra et al. 2015).

The Tamil (ISO 639-3: tam) language belongs to the Dravidian languages and is mainly spoken in Tamil Nadu, Sri Lanka, Singapore, Malaysia, and a small population in Canada. The Tamil language has official status in Tamil Nadu state, India, Sri Lanka, and Singapore. As far as we know, there is no such dataset for automatic identification of homophobia or transphobia available online. This is even worse for under-resourced languages like Tamil, where the culture itself imposes the topic of LGBT+ as taboo, and even if one of the popular social media people raises a supporting voice, they are targetted and labelled as LGBT+ (Somasundaram and Murthy 2016). The utilisation of taboo words simultaneously marks them as LGBT+ members, different or defiant by breaking accepted practices even if they are not LGBT+ and can be utilised to kick him/her out of a society which will cause severe mental health issues to vulnerable youngsters. A rare study on automatically identifying homophobic/transphobic content in social media of LGBT+ populations, particularly in the Tamil population. Our work creates a dataset for three settings, English, Tamil, and code-mixed Tamil-English.

The dataset contains some unique features that distinguish it from prior hate speech or offensive language identification datasets. First, we employ chronological and organised discussion comments, which means that annotators consider the preceding comments before labelling each post. Second, we distinguish between several sorts of LGBT+ abuse, such as derogation and personally threatening language. We also look at the counter speech against discrimination and hope speech. Third, instead of using crowd-sourced labourers, we utilise trained volunteer annotators from LGBT+ society who identify themselves as LGBT+ or LGBT+ alleys. We also employ guided meetings rather than a majority vote to settle on the final labels. These two variables combine to provide a high-quality dataset. We believe our work will automatically help to improve the detection of homophobic and transphobic comments. In addition, it can be used to either eliminate or create counter-narratives (i.e. informed textual responses).

The principal contributions of the paper are listed below:

- We perform a comprehensive and in-depth investigation of online homophobia, which is an essential and timely assignment given the increasing prevalence of hate speech and online harassment toward LGBT+ vulnerable on social media.
- We provide a new labelling taxonomy for online homophobia and transphobia, as well as an expert-labelled dataset that will allow homophobic/transphobic content to be automatically identified.
- Qualitative corpora analysis has been done.
- A cutting-edge methodology for detecting homophobia/transphobia in social media has been designed and tested in two languages English and Tamil. We also investigated if it is possible to detect homophobia in a multilingual code-mixed setting.
- We use a variety of feature extraction techniques, including TF-IDF (tri-gram), count vectorizer (tri-gram), fasText, and BERT to train numerous 3-class, 5-class, and 7-class homophobia/transphobia classifiers based on machine learning (Support Vector Machine, Naive Bayes, Logistic Regression, Random Forest) and deep learning (BiLSTM and mBERT).
- A considerable number of learnt multi-class classifiers trained using diverse feature spaces are evaluated in large-scale empirical research setup baseline for our dataset. For each classification task, we aim to identify the optimal mix of feature extraction techniques and classification algorithms so that we can adequately classify homophobic and transphobic abuses. The significance of the improvements has been shown through statistical tests.
- We conduct an error analysis of the top classifiers for each of the three language settings regarding the incorrectly predicted comments. In addition, we give complex examples with English translations for each assignment and target class. As a result, we look at a total of 27 misclassified cases.

**NOTE:** This article contains examples of language that may be offensive to some readers. They do not represent the views of the authors.

## 2. Related Work

The usage of information and communication technology, particularly social media, has altered the way people connect and create connections, as social media apps are widely used worldwide. For example, YouTube is a prominent social networking website that allows users to establish their own profiles, upload videos and leave comments. It has a large audience, as thousands may see each video or remark of people, thanks to “liking” and “sharing” methods, allowing cyberbullies to spread ugly or unwelcome information about their victims effortlessly. This unfortunately, provides an avenue for anti-social behaviors such as misogyny (Mulki and Ghanem 2021), sexism, homophobia (Diefendorf and Bridges 2020), transphobia (Giametta and Havkin 2021), and racism (Larimore et al. 2021). When it comes to crawling social media data, there are many works on YouTube mining (Marrese-Taylor et al. 2017; Muralidhar et al. 2018), mainly focused on exploiting user comments. Krishna et al. (2013) did an opinion mining and trend analysis on YouTube comments.

With the rapidly rising use of social media, computer scientists began to investigate text-based algorithms for identifying offensive languages and hate speech by crawling social media data. Multiple datasets have been created to identify hate speech (Burnap and Williams 2016), racial bias in hate speech (de Gibert et al. 2018; Davidson et al. 2019), countering hate speech (Qian et al. 2019), hierarchically labelled hate speech (Fortuna et al. 2019), abusive language (Mulki et al. 2019), bullying detection (Xu et al. 2012) and offensive language identification (Zampieri et al. 2019b; Sigurbergsson and Derczynski 2020; Çöltekin 2020).

One of the early studies on Tamil offensive language identification was conducted in 2020 by HASOC-DravidianCodeMix (Chakravarthi et al. 2020a; Mandl et al. 2020) followed by DravidianLangTech shared task (Chakravarthi et al. 2021) in which a Tamil dataset of offensive comments have been built and shared to participants of the shared task. The dataset for HASOC-DravidianCodeMix is composed of 4,000 comments that have been collected from Twitter and the Helo App. The dataset for DravidianLangTech is composed of 30,000 comments that have been collected from YouTube. Multiple annotators labelled the dataset at different levels of offensiveness. Both the dataset were code-mixed datasets. Based on a well-annotated dataset, a wide range of systems were assessed on two tasks in the Tamil language. Chakravarthi (2020) created a hope speech dataset and conducted a shared task (Chakravarthi and Muralidaran 2021) to improve the research in a positive direction for Tamil, Malayalam, and English. The authors created a Hope Speech dataset for Equality, Diversity, and Inclusion (HopeEDI) from user-generated comments on YouTube, comprising 28,451, 20,198, and 10,705 comments in English, Tamil, and Malayalam, respectively, manually classified as containing hope speech or not. This is one of the earliest datasets in Tamil which look at LGBT issues. These datasets provided an opportunity for researchers to test models on code-mixed Tamil data to stop offensive language online.

For the English language, methods to minimize gender bias in NLP have been intensively investigated (Sun et al. 2019). Machine translation to French and other languages has been used in several research works to look at gender prejudice outside of the English (Vanmassenhove et al. 2018; Prates et al. 2020). Tatman (2017) looked at the gender and dialect bias in YouTube subtitles that were created automatically. In social media, misogynistic speech is a well-studied topic, and its detection is frequently framed as a text classification challenge (Pamungkas et al. 2020). Multiple datasets were created to identify the misogyny at different levels in a different language by researchers (Parikh et al. 2019; Bhattacharya et al. 2020; Mulki and Ghanem 2021). However, homophobia or transphobia was not given much importance. Recently, an expert annotated dataset for identifying online misogyny was created by Guest et al. (2021); inspired by their work; we created our dataset.

Wu and Hsieh (2017) examine linguistic behaviour found in LGBT+ created Chinese texts and have shown that traditional systems trained to detect gender from text fail in more complex dimensions. Ljubešić et al. (2020) constructed emotion lexicons for Croatian, Dutch and Slovene and used these lexicons to find texts containing and not containing socially unacceptable discourse on the topics of migrants and LGBT+. Because the topic is still in its early stages, it suffers from several flaws connected to both the specific aims and subtleties of offensive language towards homophobia/transphobia and the nature of the classification task in general, which prohibit systems from achieving ideal results. One of the critical challenges is the inherent complexity in defining offensive language, as well as the pervasive ambiguity in the usage of similar phrases (such as abusive, poisonous, harmful, hateful, or violent language), which changes culturally and are open to highly subjective interpretations according to individuals. The significance of this essential but fundamental distinction – between sexism and androcentrism – is understandable if we use research on “sissy boys” to explain transphobia, one of the cruellest labels ever developed, which views the transgender person as suffering from a psychiatric condition. Measures of homophobia are thus influenced by how homosexuality and homophobia are seen in society and by researchers at the time the test is developed.

Our work differs from previous works in that we define the taxonomy of homophobia and transphobia at different levels, look at hope speech and counter-speech, and introduce a dataset for English, Tamil and Tamil-English code-mixed texts on the subject. To the best of our knowledge, this is the first work to create a dataset for homophobia and transphobia in English, Tamil and Tamil-English code-mixed text, where Tamil is an under-resourced language.

## 3. Data collection

Social media, such as Twitter, Facebook, and YouTube, contain rapidly changing data generated by millions of users that may radically affect an individual's or an organization's reputation. This emphasizes the importance of emotion extraction software and inappropriate language detection in online social media. Because of the enormous range of material accessible on the website, such as music, lessons, product reviews, trailers, and so on, YouTube is becoming increasingly popular throughout the Indian subcontinent. Users can create material on YouTube, and other users can comment on it. Thus, it enables more user-generated content in languages with limited resources. This is also the same for LGBT+ vulnerable who watch similar videos and comment about the video which they connect to. We chose to collect data from social media comments on YouTube [a], the most extensively used medium in the world for expressing an opinion about a particular video.

Young LGBT+ vulnerable individuals, predictably, are described as an "invisible" minority and one of the most significant "at-risk" groups of adolescents in India, a country where marriage equality is not legal. These individuals use social media as the only option to find similar people by coming out. We did not use comments from LGBT+ people's personal coming out stories since they contained personal information. Instead, we collected the videos from popular YouTubers who explain LGBT+, hoping that people will be more positive. To ensure that our dataset contains enough homophobic and transphobic abuse, we began with targeted videos of pranks from YouTubers such as "Gay Prank", "Transgender Prank" and "legalizing homosexuality". There were some videos about the positive sides of transgenderism, but most of the videos from the news channel and popular channels depict transgender people as exploiters and instigators of fights. It was challenging to find a YouTube video that talks about LGBT+ issues in Tamil since it is still taboo and marriage equality is not legal, and also, until recently, homosexuality was illegal in India. We collected all those videos.

The *YouTube Comment Scraper tool*[b] was used to collect the comments. We utilized these comments to make our datasets with manual annotations. We wanted to collect Tamil comments. However, we discovered that there was a lot of English and code-mixing throughout the text. The mixing of linguistic units from two or more languages in a single conversation or, in some cases, a single utterance is referred to as code-mixing (Pratapa et al. 2018). Code mixing is a typical, natural outcome of bilingual and multilingual language use of Tamil people. The Tamil language has a native script that has its Unicode, but still, social media users write in the Roman alphabet for convenience (phonetic typing), which enhances the chances of code-mixing with a Roman alphabet language (Jose et al. 2020; Chakravarthi et al. 2020b).

Extracting the essential content from the comment area that fitted our aim was a challenging effort, made much more difficult by the existence of responses in non-target languages. We used *langdetect library*[c] to distinguish distinct languages apart and separate languages as part of the preparation processes to clean the data. We split the data into three, such as Tamil and

[a] https://www.youtube.com/
[b] https://github.com/philbot9/youtube-remarkscraper
[c] https://pypi.org/venture/langdetect/

English based *langdetect library*. We kept the remaining code-mixed Tamil-English. We eliminated all user-related information from the corpus in line with data privacy regulations. We deleted extraneous information, such as URLs, as part of the test preparation.

Because we gathered the corpora from social media, they comprise a variety of real-world code-mixed data even though *langdect* classified them as Tamil or English. For the code-mixed data, it is another case. Inter-sentential mixing is defined as a shift in language between sentences in which each phrase is written or spoken in a different language. Intra-sentential mixing happens within a single phrase; intra-sentential mixing happens when one phrase is in one language and the other is in a different language. Inter-sentential mix, intra-sentential mix, and tag switching are the three forms of code-mixed phrases we encountered. Most of the remarks were written in Roman character and used either Tamil grammar or English grammar with Tamil vocabulary. All the comments were collected between August 2020 and Feb 2021.

**Table 1.** Raw dataset statistics by language based on langdect

| Language | Number of comments | Number of tokens | Number of Characters |
|---|---:|---:|---:|
| English | 7,265 | 116,015 | 632,221 |
| Tamil | 5,240 | 255,578 | 787,177 |
| Tamil-English | 10,319 | 88,303 | 628,077 |
| Total | 22,824 | 249,896 | 2,047,475 |

## 4. Taxonomy

Homophobia and transphobia are two concepts that relate to unfavourable views toward homosexual and transsexual persons (Haaga 1991). Transphobia is a severe problem that affects many individuals. It is described as fear and/or hate of transgender persons. In both gay and straight cultures, transgender persons are frequently shunned and neglected. Many transgender persons are prevented from coming out or identifying themselves due to ignorance and hostility, further obscuring them. We developed a hierarchical taxonomy with two levels. First, we make a ternary distinction between homophobic content, transphobic content and non-anti-LGBT+ content.

We elaborated subtypes of homophobic, transphobic, and non-anti-LGBT+ content. For homophobic and transphobic content, we defined two categories: (i) Homophobic/Transphobic Derogation and (ii) personal attacks against LGBT+ people called Homophobic/Transphobic Threatening. For Non-anti-LGBT+, we defined three categories: (i) Counter speech against homophobia/transphobia, (ii) Hope Speech and (iii) None of the categories. This taxonomy is inspired based on the misogyny typologies offered by Guest et al. (2021)

### *4.1 Homophobic content*

Homophobic content can be described as a type of gender-based harassment comment that involves the use of pejorative labels (e.g., ‘fag’, ‘homo’) or denigrative phrases (e.g., ‘don’t be a homo’, ‘that’s so gay’) directed against people who are gay, lesbian, bisexual, queer, or gender non-conforming (Meyer 2008; Poteat and Rivers 2010). The most common definition of homophobia is “an attitude of hostility toward male or female homosexuals” (Fraïssé and Barrientos 2016). However, there is a contrast between general and specialized homophobia and between families of phobias targeting various target groups: lesbophobia, gayphobia, and biphobia. In this paper, we handle lesbophobia, gayphobia, and biphobia under homophobia.

#### 4.1.1 Homophobic pejoratives or derogation

Homophobic pejoratives are phrases used to denigrate a low opinion or a lack of respect toward the LGBIQ vulnerable. It can also be expressing contempt or disapproval for LGBIQ vulnerable, essentially more subtle yet effectively homophobic. It includes phrases that are expressly offensive and disparaging, such as 'homo,' 'fag,' 'dyke' and 'lezza' as well as expressions that communicate implicit hostility or anger against LGBIQ, such as 'no homo.' Similarly, the terms 'gay' and 'poof' are frequently used to describe anything judged unmasculine, non-normative, or 'uncool' (Thurlow 2001). Example phrases are 'that's so gay', 'you're so gay', and 'you're so lesbian'.

Derogatory content about LGBIQ is known as homophobic derogation. This content may be abusive, either directly or indirectly. Unfavourable stereotypes about homosexual men's manly skills, such as a lack of physical strength or emotional control. They were expressing unfavourable moral judgments about gay men, such as implying that they are weak or inferior to males in some way. Creating unfavorable assumptions about the LGBIQ relationship and/or sexual abilities includes perceived unattractiveness (a lack of sexual appeal), ugliness (a lack of beauty), frigidity (a lack of sexual readiness), and demeaning assertions about LGBIQ capacity to live long mates. Examples would be 'Dont cry like fag or gay', 'All gays/lesbians/bisexuals are promiscuous, 'Gays are always camp', 'All lesbians are bull dykes', 'Gays sleep with people for money, or 'Gays want to sleep with every guy in the room'. For our annotation we keep both pejorative and derogatory as Homophobic derogation.

#### 4.1.2 Homophobic threatening language

Content implies an intent/desire to hurt/cause harm to LGB, or supports, promotes, advocates, or incites such violence. It is a type of abuse that's "explicit." It can be physical non sexual acts of violence such as killing, maiming, beating, or explicit sexual violence such as rape, molestation or penetration or an invasion of privacy such the disclosure of personal information. Example phrases include "Gays deserve to be shot dead", "Someone should rape that lesbo to turn her into straight", "Gays should be stoned", "You lesbos I know where you live, I will visit you tonight", " beat the fag out of him", "You should kill yourself".

### 4.2 Transphobic content

The reader might ask why not keep both homophobia and transphobia under the same label, but there are subtle differences. Despite their similarities, transphobia and homophobia are not the same thing. Transgender people identify with a gender other than the one assigned to them at birth (Graham et al. 2011). A heterosexual person, for example, is someone who was assigned female at birth but now identifies as a male. Today, many transgender people use sexual orientation terminology to refer to their present gender identification rather than their gender at birth. Because many regulations meant to protect LGB persons do not include protections based on gender identity or expression, transgender adolescents experience much greater marginalization and have less access to remedies than their LGB peers in some parts of the world (McGuire et al. 2010; Hatchel et al. 2019). Transphobic persons can be homosexual or straight, and they can be transphobic without being homophobic. Transgender people has a legal status in India, and they have to are given special status due to mythological figures and their association with Hindu gods. However, LGBQ people do not have the right to marry in India. Homophobia is taboo, not even spoken in Tamil Nadu.

#### *4.2.1 Transphobic pejoratives or derogation*

Transphobic pejoratives are phrases that are used to disparage vulnerable transgender people. It includes phrases that are expressly offensive and disparaging, such as ‘tranny’, ‘trannie’, ‘cross-dresser’, or ‘drag’ as well as expressions that communicate implicit hostility or anger against transgender, such as ‘She-Male’, ‘it’, ‘9.’ Similarly, the terms “not man enough”, “not women enough”, “will never be full man”, and “will never be full women” are frequently used to describe. Examples include “Trans people are confused people who know no better” and “There are huge links to autism and transsexuality, so all trans are autistic”.

Derogatory content about transgender people is known as transphobic derogation. This content may be abusive, either directly or indirectly. Unfavorable stereotypes about vulnerable transgender peoples skills, such as a lack of womenelinss or emotional control. Expressing unfavorable moral judgments about transgender people, such as implying that they are not equal to or inferior to women or males in some way. An example would be ‘All Transgenders women are whores’. Making negative judgements of transgender’s relationship and/or sexual ability. This includes perceived unattractiveness (i.e. a lack of sexual desirability), ugliness (i.e. a lack of beauty), as well as belittling statements about trangender peoples ability to find life time partners. For our annotation we keep both pejorative and derogatory as transphobic derogation.

#### *4.2.2 Transphobic threatening language*

Transphobic threatening language content is defined as anything that discusses, advocates, incites, or plans unfavorable or detrimental treatment of transgender. It covers both stating intent to take action against transgender and expressing preferences for how they should be treated which includes threatening language, physical violence, sexual violence, or invasion of privacy. Examples phrases includes ‘Kill those tranny no one will care’, ‘trannie deserve to be beaten up’, and ‘They asked for a beating when they crossdress’. Asking them to kill themselves is also threatening

### *4.3 Non-anti-LGBT+ content*

Non-anti-LGBT+ information can be classified into three groups, all of which are important to homophobia and transphobia study. First, Wright and Wachs (2021) showed that toxic online disinhibition and poor empathy are associated with homophobic/transphobic cyberbullying. Second, by studying non-anti-LGBT+ comments, we can help design preventative and intervention programs targeted at altering social media user online actions and views.

#### *4.3.1 Counter Speech*

Counter speech is content that criticizes, refutes, or calls past homophobic or transphobic abuse in a comment into question (Tekiroğlu et al. 2020; Garland et al. 2020). More specifically, counter speech is a non-aggressive reaction that provides feedback through fact-based arguments. For example, it might criticize past abuse explicitly (e.g., ‘What you said is unacceptable’), expressly accuse it of bias (e.g., ‘That’s terribly homophobic’), or present a contrasting viewpoint that contradicts the homophobia/transphobia (e.g., ‘That’s not how LGBT+ act, you’re so incorrect’).

#### *4.3.2 Hope Speech*

Hope speech engenders optimism and resilience that positively influences the readers (Youssef and Luthans 2007). We define hope speech as inspirational talk about how people face difficult situations and survive them (Snyder et al. 2002). The comment contains inspiration provided to participants by their peers and others and/or offers support, reassurance, suggestions and insight (Chakravarthi 2020; Chakravarthi and Muralidaran 2021). Empathetic responses are also considered as hope speech. For example, “It was very hard in school days, now in college, it is better, hang on it will get better”.

#### 4.3.3 None of categories

Content does not include homophobic or transphobic insults, pejoratives, or counter-speech as outlined in the preceding categories. This topic is frequently unrelated to abuse or vulnerable LGBT+ people in general. Such as asking for liking the video or subscribe to the channel, or asking to like the comment. However, other sorts of offensive language that are not anti-LGBT+ can be included.

## 5. Annotation

We carefully examine the text once it has been preprocessed and cleaned to ensure that it contains no personal information. Producing high-quality annotations is a significant challenge when creating training datasets for a sensitive topics like ours. Several factors impact this including annotators cultural influences and personal biases. Since the LGBT+ topic is a taboo topic, it was challenging to find annotators. For English, we used in-house LGBT+ societies members. It was challenging for Tamil to find annotators who are willing to annotate; even Tamil people in culturally progressive places like Europe, United Kingdom, and the United States did not want to be associated with this work, thinking that they might be assumed as LGBT+. After a long search, we found LGBT+ or LGBT+ ally volunteer annotators, and we gave training by showing YouTube videos explaining what is LGBT+ means [d] and then gave a video which explains whether LGBT+ is normal or abnormal [e]. The annotation interface displayed the current sentence, as well as the previous and future sentences.

### 5.1 Ethics

We are dedicated to adhering to ethical standards, which includes safeguarding vulnerable individuals privacy and confidentiality. We removed the user ids, phone number or address given in the comment before sending it to annotators. We removed the user ids, and phone numbers automatically. Address was removed manually. Data from social media is extremely sensitive, especially when it concerns minorities like the LGBTIQ community. By deleting personal information from the dataset, such as names but not celebrity names, we have taken great care to minimize the danger of individual identity in the data. However, in order to research homophobia/transphobia, we needed to keep track of information on sexual orientation, gender and philosophical ideas. Annotators have only viewed anonymized postings and promised not to contact the author of the remark. Only researchers who agree to follow ethical criteria will be given access to the dataset for research purposes. The annotators were given a choice to quit the annotation whenever they are uncomfortable with annotation.

### 5.2 Annotation Process

Instructions and mentoring were offered to annotators on how to classify social media material for labels explained above. We ask the annotators to consider a comment a positive example of homophobia/transphobia if it conveys similar examples to that of the above taxonomy examples. We used Google Forms to obtain the annotator’s email address so that they may only annotate once. A maximum of 100 comments is allowed on each Google Form. We distributed the Google form to members of the LGBT society of the National University of Ireland Galway for English, for Tamil and code-mixed Tamil-English we sent it to many LGBT+ societies in Tamil Nadu, but we got very few responses. Seven annotators offered to annotate in the end for Tamil and Tamil-English code mixed. All of them are Tamil-English bilinguals and prepared to take on the assignment seriously. For English, we got responses from 3 members. All of the annotators were graduates or postgraduates since we contacted people from college societies. All the annotators identified themselves as LGBT+ or LGBT ally.

[d]https://www.youtube.com/watch?v=Tz-DsgYHF1A
[e]https://www.youtube.com/watch?v=pSmZCS31LWc

**Table 2.** Annotators Statistics by language

| Language | | English | Tamil | Tamil-English |
|---|---|---|---|---|
| Gender | Male | 1 | 3 | 2 |
| | Female | 1 | 2 | 1 |
| | Non-binary | 1 | 0 | 0 |
| Higher Education | Undegraduate | 0 | 2 | 0 |
| | Graduate | 0 | 0 | 1 |
| | Postgraduate | 3 | 3 | 2 |
| Medium of Schooling | English | 3 | 1 | 3 |
| | Tamil | 0 | 4 | 0 |
| Total | | 3 | 4 | 3 |

**Table 3.** D ataset statistics after annotation

| Language | Number of comments | Number of tokens | Number of Characters |
|---|---|---|---|
| English | 4,946 | 82,111 | 438,980 |
| Tamil | 4,161 | 197,237 | 539,559 |
| Tamil-English | 6,034 | 66,731 | 435,890 |
| Total | 15,141 | 346,079 | 1,414,429 |

It was not easy to find annotators for Tamil. More than 20 annotators dropped after two forms due to it being a sensitive topic, and it took a long time to annotate for them because they did not understand these issues. We have given them 30 min training explaining these terms along with the YouTube videos from popular YouTuber explaining about LGBT+ people and their issues. A minimum of three annotators annotated each comment. If more than three annotators agreed, then we took them as labels otherwise flagged as disagreements. Annotators and the along with the authors of this paper reviewed the disagreement overseen by the first author, who had developed the annotation taxonomy and was familiar with the literature on homophobia and transphobia, and who is also an expert in both the Tamil and English language. The facilitator was to encourage discussion among annotators and ensure that the final labels followed the taxonomy. The discussion happened online via Google Meet. Each point of disagreement was addressed until the annotators agreed on the final label. If no agreement could be concluded, those comments were discarded from the dataset for this study. For Indian annotators, we paid 300 rupees per hour; the median salary in India is around Rs. 16,000 per month (Rs. 533 per day) according to moneymint.com[f].

[f]https://moneymint.com/what-is-average-salary-in-india/

Details about the annotator are given in Table 2. From Table 2, we can see that we have fewer number of final annotators even though we got 20 volunteers in the beginning. Most of the annotators were postgraduate, except in the case of Tamil. Tamil comments were annotated by Tamil people who studied their schooling in the Tamil. However, for Tamil-English we needed people who are proficient in both Tamil and English, so we gave the forms to Tamil people who have both English and Tamil language speaking, reading and writing proficiency. There was an option to choose non-English or non-Tamil in our forms. Even though the *langdect* library was suitable for detecting the sentence's languages, they did not always detect it correctly. We got more non-English comments in the English only comments annotation process. Since it was annotated by volunteers English native speakers from the Republic of Ireland, they marked the code-mixed Tamil-English as non-English. All non-intended language comments were removed.

### 5.3 Annotation Evaluation

The degree to which annotators agree in their ratings is measured by inter-annotator agreement. This is required to guarantee that the annotation system is consistent and that several raters can assign the same sentiment label to the same comment. We used the Krippendorff's alpha coefficient (Krippendorff 1970). We use Krippendorff's alpha because it can be used with any number of annotators, each assigning one value to one unit of analysis, missing (incomplete) data, any number of values available for coding a variable, binary, nominal, ordinal, interval, ratio, polar, and circular metrics, as well as binary, nominal, ordinal, interval, ratio, polar, and circular metrics (levels of measurement). $\alpha$ 's general form is:

$$\alpha = 1 - \frac{D_o}{D_e} \tag{1}$$

where $D_o$ is the observed disagreement among values assigned to units of analysis:

$$D_o = \frac{1}{n} \sum_c \sum_k o_{ck} \, {}_{\text{metric}}\delta^2_{ck} \tag{2}$$

and $D_e$ is the disagreement one would expect when the coding of units is attributable to chance rather than to the properties of these units:

$$D_e = \frac{1}{n(n-1)} \sum_c \sum_k n_c \cdot n_k \, {}_{\text{metric}}\delta^2_{ck} \tag{3}$$

When annotators agree exactly, observed disagreement $D_o = 0$ and $\alpha = 1$, which indicates perfect reliability among annotators. When observers agree as if chance had produced the results, $D_o = D_e$ and $\alpha = 0$, which indicates the absence of reliability. Finally, $\alpha = 0$ occurs when annotators are unable to come to an agreement. To rely on data created by any method, $\alpha$ needs to be far from these two extreme conditions, ideally $\alpha = 1$. For reliability considerations, $\alpha$ 's range is:

$$1 \geq \alpha \geq 0 \begin{cases} - \text{ Systematic disagreement} \\ \pm \text{ Sampling errors} \end{cases} \tag{4}$$

The agreement coefficient calculates the amount that annotators agreed on label assignments over what would be anticipated by chance. We used NLTK's nltk.metrics.agreement module[g] to calculate the $\alpha$. We got Krippendorf's alpha of 0.67, 0.76, and 0.54 for English, Tamil and Tamil-English respectively.

[g]https://www.nltk.org/

**Table 4.** Dataset distribution at first level

| | English | Tamil | Tamil-English |
|---|---|---|---|
| Homophobic | 276 | 723 | 465 |
| Transphobic | 13 | 233 | 184 |
| Non-anti-LGBT+ content | 4,657 | 3,205 | 5,385 |
| Total | 4,946 | 4,161 | 6,034 |

**Table 5.** Dataset distribution at second level

| | English | Tamil | Tamil-English |
|---|---|---|---|
| Homophobic | 276 | 723 | 465 |
| Transphobic | 13 | 233 | 184 |
| Counter speech | 486 | 336 | 281 |
| Hope speech | 687 | 335 | 317 |
| None of the above | 3,484 | 2,534 | 4,787 |
| Total | 4,946 | 4,161 | 6,034 |

**Table 6.** Dataset distribution

| | English | Tamil | Tamil-English |
|---|---|---|---|
| Homophobic-derogation | 262 | 661 | 408 |
| Homophobic-threatening | 14 | 62 | 57 |
| Transphobic-derogation | 11 | 170 | 105 |
| Transphobic-threatening | 2 | 63 | 79 |
| Counter speech | 486 | 336 | 281 |
| Hope speech | 687 | 335 | 317 |
| None of the above | 3,484 | 2,534 | 4,787 |
| Total | 4,946 | 4,161 | 6,034 |

### *5.4 Difficulties in Annotation*

The comments were code mixed and written in the Roman script without proper spelling or no standard spelling; since Tamil is diglossic, the dialects come with different spellings, making it even more difficult for the annotator to understand some comments. In addition to that, some of the words are mixing of English-like words but not English, so the annotators had to read the comments multiple times. For example, pool (‘dick’) or poola(‘like’) are confused for English words or English words with Tamil morphology but they are Tamil words. Particularly the word ‘sissy’ is used as ‘sister’ in Tamil code-mixed and English comments from our dataset. However, instead of the derogatory meaning of ‘sissy’ it was used in a polite way, and the number ‘9’ was used to insult LGBT+, so we had to explain this to Irish annotators who annotated English comments since it is a cultural difference.

## 6. Corpus Analysis

We collected as much data as possible for an annotation since this topic is taboo and there is not much content available in Tamil on YouTube. Statistics of the raw data, including the number of comments, tokens and characters in our data by each language setting, is shown in Table 1. Table 3 shows the data statistics after the annotation, the reduction of the comments are due to not being in the intended language as well the comments where the annotators disagreed. In total, our dataset contains 15,141 comments from YouTube videos, where 4,946 comments are in English, 4,161 comments in Tamil, and 6,034 comments in Tamil-English. We used the 'wc' tool in Linux to tokenize and calculate the number of comments. As shown, the number of tokens in Tamil is high due to the Tamil script.

Table 4, Table 5, and Table 6 presents the distribution of the annotated dataset by the label at three-level, five level and seven level hierarchy. A detailed distribution of the annotated dataset is shown in Table 6. The dataset is skewed, with almost the majority of the comments being labelled as non-anti-LGBT+ content for all three language settings. This is common for user-generated content on online platforms, and an automatic detection systems need to handle imbalanced data. Classwise, our dataset contains more homophobic comments than transphobic comments. In English, the homophobic and transphobic percentage is low; this might be due to automatic deletion or blocking of such comments by YouTube. However, this is not the case for Tamil or Tamil-English. For the fine-grained classification, the dataset contains very few comments, such as only two comments annotated as transphobic-threatening and 14 for homophobic-threatening in the English language. However, we see threatening is above 50 for the Tamil and Tamil-English code-mixed dataset. We also see many hope speech comments in English compared to Tamil and Tamil-English. Counter speech is more or less similar in all three settings. Out of derogation and threatening comments, the homophobic derogation is high in all three language settings. We have given examples of annotated comments for each class in all three language settings.

### *6.1 Examples of annotated dataset for each class*

#### *6.1.1 Homophobic-derogation*

- தேவிடியா பையா... இன்றைக்கு பொட்டைபசங்க காதல் வைரக்கும் படம் எடுத்தவன் நாளைக்கு அம்மாவுக்கும் பையனுக்கும், அப்பாவுக்கும் பொண்ணுக்கும்ன்னு எடுக்க கூட தயங்கமாட்ட டா நாயே... உங்க அம்மா அப்பா செய்த பாவம்.... - *Prostitute boys... today they take love story of fag boys, tomorrow he will make love story of son and mother or son and father dogs... it is your parents faults that you are born...* This comment is disturbingly homophobic derogation.
- ஓரினச்சேர்க்கை தவறு இல்லைனா அப்பன் மகனோடு உறவு கொள்ளலாமா..?; அண்ணன் தம்பியோடும் தம்பி பெரியப்பனோடும் உறவு கொள்ளலாமா..?- *If Homosexual is not wrong then can you have sex with your parents?, can siblings have sex with uncles?* . This comment is homophobic deragation since it compares the LGB relationship to incest sexual relationship.
- **Iva naaya vida kevalamaanaval**- *She is worse than the dog*. This is compare the lesbian with dog and making worse than dog, so this is homophobic comment.
- **Appo thappu illana ni oru naikittayo illa pannikkittayo sex vachikko** - *If it is not wrong then go and have sex with dog and pigs*. This comment is also homophobic derogation which directly compares to animal sex with LGB+ peoples.
- **This is actually Disease** This comment is homophobic derogation since it compares homosexuality to a disease.
- **There is a saying EVE CREATE FOR ADAM „„„BUT NOT STEVE CREATE ADAM** This comment is also homophobic derogation since it uses bible mythology against LGB+ people.

### *6.1.2 Homophobic-threatening*

- திருட்டு நாய்கள் ஓரினச்சேர்க்கையில் ஈடுபடுபவர்களை அடித்து கொலை செய்ய வேண்டும் - *Thieving dogs, we should beat homosexuals to death.* This comment initiates death threat against the LGB+ peoples.
- இவளையெல்லாம் நாட்டில விட்டு வைப்பதே மிக தப்பு நாய் - *We should not let lesbians live in our country* . This comment is violent and endangers the basic right of the LGB+ people.
- இந்த மாதிரி நாய்களை சுட்டு கொள்ளவேண்டும் - *Dogs like this should be shot.* This comment initiates a death threat.
- **avana nee parathesi poai saavuda** - *Are you one of homosexual homeless f--ker, you should kill yourself.* This comment is also homophobic threatening since it is asking LGB+ person to commit suicide.
- **Madam read bible ; What bible says about homosex ; Jesus destroy one country ; For homosex ; ; Jesus coming soon to destroy your country**. This comment is threatening LGB+ people using Biblical mythology.

### *6.1.3 Transphobic derogation*

- அரவானிகள் சமூக நாசக்காரர்கள்,.. - *Transgender are destroyer of community (aravani is derogated word for Transgender).* The respectable word for Transgender in Tamil is 'thirunangai' however the author used 'aravani' and said that the transgender destroy community so this comment is transphobic derogation.
- அரவானிகள் யதார்த்தமாகவே நல்லவர்கள் கிடையாது, இதுதான் அவர்கள் சுயரூபம்,.. - *Transgender are inherently not good, this is their nature,.* This comment also used 'aravani' word instead of 'thirunangai' and generally attack that all transgenders are bad.
- **io samy onbathukitta maatuna summa vituvangala** - *Oh my god, if you get caught by transgender then you cannot escape*. This comment is transphobic derogation since it potrays all transgender as bad people.
- **Appo transgender ellaam ugly** -*All transgenders are ugly* . This is also portrays transgenders in bad way.
- **Very ugly stupid thirunangai** This comment is also transphobic derogation since it targets transgender and saying they are ugly.
- **Body is female... But brain is male... Genetic disorder** This comment is transphobic derogation since it compares it to genetic disorders and demeans transgenders.

### *6.1.4 Transphobic-threatening*

- இந்த நாய்களை என்கவுண்டரில் போடனும் - *We should kill these transgender in police encounter* This comment is transphobic threatening since it is asking the police to kill transgender and get away without any case.
- இந்த நாய்களை பார்த்தவுடன் சுட்டு தள்ள உத்தரவு செய்ய வேண்டும். - *We have to make an order to shot these dogs dead on site.* This comment is also initiate a violence against vulnerable transgender so this is transphobic threatening.
- **Sava adiga indha nayegalae thirunagai bastard's** - *Kill these transgender bastard* . This comment is also transphobic threatening since they are asking to kill transgender people.
- **Easiest way is to just stab them in the streets. Or We can poison their water supply. ; But you are just sitting behind a keyboard and a useless piece of shit. So I can't count on you. ; I will look for others who can actually kill them. And cleanse the country.** This comment is planning to murder transgenders and asking people to join him/her so this is also transphobic threatening.

*6.1.5 Counter speech*

- மற்றவர்களின் உணர்ச்சிகைளயும் மதிப்ேபாம் - *We should respect others feeling*. This comment is counter speech since it is asking the people to respect everyone feelings.
- தனி ஒரு திருநங்ைக தவறு ெசய்தால் அைனத்து திருநங்ைக ைளயும் ெபாதுைமபடுத்தி ேபசுவது சரியல்ல... - *Because one individual transgender did a bad thing you cannot blame all transgender in general*. This comment is counter speech since it opposes the transphobia and argues that we cannot generalizes due to one individual.
- ஐயப்பன் பிறப்ைப ஆதரிக்கின்றவர்கள் ஏன் இைத ெகாச்ைச படுத்துகிறார்கள் - *You can accept the birth of God Iyyappan then why are homophobic*. This comment is counter speech since it uses the God Ayyappan[h] who is a son of two male principal deities (Shiva[i] and Thirumal[j] of Hinduism to say that you already accepted these people as god then why discriminate.
- **Arthanarishwar Kalla Irundha Poyi Kaiya Eduthu Kumbuduringa...Adhuve Manushana Porandha Evlo Cheap ah Nadathuringa...Avangellam Manusha Uruvathula Irukka Kadavul Thirunangaiya Irundhaave Thappa dha Nenaikiringa...Namma dha Avangala Thappana Vazhikku Poga Innum Force pannittu Irukkom...Ckram Ellaame Maarum...Maathuvom; Manasula Irundhu Solra Neenga Ellaarum Nalla Iruppinga; Innum Indha Maadhiri Nalla Videos pannunga...Much Love** - *You pray to the God Arthanarishwar[k] even though it is stone, but if it was a human you treat them bad. They are also God in the form of human. We are the ones who force them to bad lifestyle by not accepting them into our normal society. We should change and we shold accept them into our society so they can also live life with pride. Much love.* . This comment is counter speech since it provides feedback through fact-based argument and using the common religious beliefs to change to opinion to positive side.
- **All had there freedom ,love had no gender don't hurt them**. This comment is counter speech by saying to not to hurt anyone.

*6.1.6 Hope Speech*

- இைறவன் உங்கள் கூட இருப்பார்... - *God will be with you*. This comment is hope speech since it is supporting and giving hope to reader.
- ஓரின ேசர்க்ைக இயற்ைகயானது - *Homosexuals are natural*. This is a hope speech since it support the LGBT+ people and gives hope for LGBT+ individuals.
- இது இயற்ைகயில் உள்ளேத அவர்கள் மனிதர்கள் தான் ; அவர்கள் காதல் மட்டும் உண்ைம - Homosexuals are in nature, they are also humans. Their love is true love. This is also hope speech since it gives hope to LGBT+ vulnerables.
- **God bless you sister jesus unkalai nechikkirar, naam anaivarum avar kulanthaigal**. - *God bless you sister, Jesus always thinks about you, we all are his children*. This comment gives hope to all by saying all are children of god and god is with them, this gives hope to vulnurale individuals.
- **love is equal for every human... love doesn't need a gender**. This comment also gives hope to LGBT+ vulnuerables.
- **Am also gay bro ; I love your videos.** This comment is also hope speech since it give hope of other LGBT+ indivduals to see that they are not alone.

[h] https://en.wikipedia.org/wiki/Ayyappan
[i] https://en.wikipedia.org/wiki/Shiva
[j] https://en.wikipedia.org/wiki/Thirumal
[k] https://en.wikipedia.org/wiki/Ardhanarishvara

#### *6.1.7 None of the above*

- நம்பைரஅனுப்புதும் பதில்இல்ைலேய.நம்பேர - *You did not reply even after I gave mynumber*. This comment is none of the above.
- சிரிப்பு காட்டாத பா.. - *Dont make me laugh*. This comment is also none of the above since it is general comment.
- நன்றி நன்றான ைஉர - *Thank you, this is good video*. This comment is also none of the above since it is general note.
- ேபாடா லூசு புண்ைட - *You mental v--na*. This is also general offensive comment it wasnot directed to LGBT+ so it was also labelled as none of the above.
- **I am waiting bro**. This is also general comment so it is labelled as none of the above.
- **The beach locations are awesome....where is it exactly?** This is also general comment so it is labelled as none of the above.

### *6.2 Qualitative Analysis*

Due to cultural and societal prejudices, annotating examples for homophobia/transphobia is a unique and challenging endeavour. Therefore, we did qualitative research to learn more about how these annotations are made and, in particular, to determine whether there are any anomalies in the data that might affect data quality.

We found that most of the counter speech is against British rule in India, blaming the British government for bringing the law to illegalise homosexuality and trying to convince the people that Tamil and Indian culture supported homosexuality and did not punish LGBT+ people. Since most of the homophobic comments claim that LGBT+ people are against nature, the counter speech tries to convince the people that homosexuality is in nature by giving the example of the animal kingdom and explaining that we should respect everyone’s feelings. Overwhelmingly, the homophobic comments target the sexual part of LGBT+ people rather than the relationship, so the counter speech also asks people to look at relationships, not just sex.

In some of the comments, the people were worried that the population of Tamil people might go down because of homosexuality, so the counter speech was explaining about adoption, surrogacy and also saying that if heterosexual people cannot have children, we would not stop them from marrying. Some of the comments blamed the Indian government for enforcing forced contraception of Tamil women and Tamil men by removing their reproductive organs after one or two children. This was in the name of population control; however Indian government did not enforce this strictly in other parts of India; even though Indian law is common to all Indian citizens, they forced only Tamil people to control the Tamil people population and so committed genocide against the Tamils. By this, they are saying population control has nothing to do with LGBT+ comments and asking the people to fight for their rights with the Indian government about the right to have children instead of taking the rights of LGBT+.

People used Tamil identity both against homosexuality and for homosexuality. For example, people would comment that this is not in our Tamil culture and blame Western people for introducing homosexuality to Tamils. While other people were commenting, saying that this was in nature and no Westerners introduced this to our culture, our culture accepts and unity in diversity, not monoculture. Most of the counter speech contains phrases like we are Tamil people and we are not Aryans, so we should accept and respect everyone’s feelings. There were so many Tamil poets who are LGBT+ people; for example, Thirumangai Alvar is a male poet who falls in love with Thirumal, a male god and many poems about his relationship with the male God.

In Tamil Nadu and among Tamil diaspora population, Tamil people follow one of the major eight prominent religions such as Tamil Shivam [l], Tamil Vainavam[m], Tamil Aseevagam[n], Tamil Jain[o], Christianity[p], Islam[q], Bhuddism[r] and Brahmanism [s] popularly called Hinduism. Out of these religions, Tamil Shivam, Tamil Vainavam, and Tamil Aseevagam are more inclusive, supportive of all life forms and counter-speech also reflected by bringing God and mythological stories from these religions. These religions were not used against the LGBT+ people in our corpus, not even in a single comment. However, Brahmanism, also popularly known as Hinduism and Islamic religion, was used by people to create homophobic and transphobic both derogation and threatening comments. We did not find any comments, either positive or negative, using Tamil Jain or Buddhism. However, Christianity was used both positively and negatively by the social media people for homophobic and transphobic comments.

Even though Tamil Shivam, Tamil Vainavam, Tamil Aseevagam, Tamil Jain and Brahmanism are put together as single religion as Hinduism outside of Tamil Nadu, – Tamil Shivam, Tamil Vainavam, Tamil Aseevagam, Tamil Jain does not want to be called Hinduism since, in Hinduism, only Brahmins can become priest and Hinduism is dominated by Brahmins. In so-called Hinduism, Brahmins are in a high position and discriminate against people based on the colour of skin and birth. While Tamil Shivam, Tamil Vainavam, Tamil Aseevagam, and Tamil Jain are more inclusive, anyone can become a priest, everyone is equal in the eyes of God, and not discriminate based on the colour of the skin. While Brahminism (Hinduism) main religious books such as Bhagavat Gita and Manusmriti calls all Tamil people except Tamil Brahmins as Shudras ("unclean, untouchable and born from feet of God saying they are worthless")[t] even though it is illegal to discriminate based on caste according to Indian law, but still Tamil people are not allowed inside Tamil Nadu Hindu temples to touch the Tamil gods or to become priests since Tamils are considered as Shudras and are not permitted to perform the upanayana, the initiatory rite into the study of the sacred literature [u] so they do not want be to called as Hindus. We see many comments reflecting the idea of the above statement when the comment creates call themselves as Tamil Shivam, Tamil Vainavam, and Tamil Aseevagam and argues why they want to be called as such. We also saw people fighting over these issues in our comments, saying they do not want to be called Hindus. There were also comments arguing about allowing Tamil language to be a religious language for practices ritual in Tamil, while some radical Hindus (Brahmanism followers) call Tamil as 'neesha bhasha'(wicked language) branded it as unfit even for worshipping Gods made it even worse. So majority of Tamil people does not want to be associated with Hinduism (Brahminism) altogether. We see these kinds of comments unrelated to homophobia and transphobia, but these are other kinds of discrimination against Tamils. These discrimination comments against Tamils were predominantly written in English or Tamil in the Roman script (Tamil-English) code-mixed, which are labelled as 'none-of-the-above' since they are not related to transphobia or homophobia.

We found the most hope speech comments are associated with God by saying God will be with you; God is with you, God loves you or showing an example of a mythological figure being LGBT+. Some of the example stories are: During the incognito (living in disguise for a year) exile period in the Mahabharata (classical war epic), Arjuna, the third of the Pandava brothers, assumed the appearance of Brihannala - a transgender person. Arjuna taught Uttara, the daughter of King

[l]https://en.wikipedia.org/wiki/Shaivism
[m]https://en.wikipedia.org/wiki/Sri_Vaishnavism
[n]https://slidetodoc.com/aseevagam-or-ajivika-aaceevakam-yennum-thamizhar-anuviyam-aaceevakam/
[o]https://en.wikipedia.org/wiki/Tamil_Jain
[p]https://en.wikipedia.org/wiki/Christianity
[q]https://en.wikipedia.org/wiki/Islam
[r]https://en.wikipedia.org/wiki/Buddhism
[s]https://www.britannica.com/topic/Brahmanism
[t]https://www.bbc.com/news/world-asia-india-35650616
[u]https://www.britannica.com/topic/Shudra

Virata, dancing and singing as Brihannala [v]. Aravanis[w] is son of Pandava prince Arjuna. Before he died, Aravan desired that he be married. Thirumal fulfilled this boon in his feminine form, Mohini (Transgender). In addition, he is the patron god of well-known transgender communities in Tamil Nadu. Ayyappan was the son of two male gods Shiva and Thirumal. Arthanarisvarar is both a male and female form of Shiva. All these stories and given as an example show the LBGT+ people to hold on to their lives and give them hope.

One of the prominent life stories of transgender was given to give inspirations to youngsters as well. Narthaki Nataraj is a transgender Bharatanatyam dancer from Tamil Nadu. In 2019, she was awarded the Padma Shri, making her the first transgender woman to be awarded India's fourth-highest civilian award. In 1984, she began studying under K. P. Kittappa Pillai, a direct descendant of the Thanjavur Quartet. Appointment as a member of the Tamil Nadu Advisory Committee in 2021. The School Education Department has incorporated a lesson on the achievements of famous Bharathanatyam dancer Narthaki Nataraj in the Tamil textbook in an effort to raise awareness about the transgender population among young minds. Her story came in multiple comments to give hope to people.

In terms of political parties, Naam Tamilar Katchi (literally meaning:"We Tamils Party"; abbrevation: NTK) is a Tamil nationalist party in the Indian states of Tamil Nadu and Puducherry, and Dravida Munnetra Kazhagam (DMK; translation. Dravidian Progressive Federation) is an Indian political party with a significant presence in Tamil Nadu and Puducherry. Both these party names were used to support LGBT+ people and gave multiple counter speech and hope speech by writing "we NTK or DMK support the rights of sexual minorities". While people blamed Bharatiya Janata Party (BJP) for legalizing LGBT+ and asking the party members to hold on to traditional values of Hinduism. The other major parties, like Congress and AIADMK parties, were not mentioned in the comments. Above study are based on social media comment from the public, it is not authors opinion about any individual, party, community or religion.

## 7. Experiment Setup

To evaluate our dataset, baseline models were built. The process of building these baselines is explained in detail below.

### *7.1 Data Preprocessing*

We have collected three corpora containing monolingual texts in Tamil and English languages and multilingual text in Tamil-English code-mixed form. Text in the corpus is full of noise since it is from social media. So, the comments on YouTube had punctuation, tags, and symbols such as @ and emojis. The pre-processing techniques, namely removal of punctuation, stop words and tags, were applied to clean the data. We divided the corpora using a stratified sampling strategy with K-folds to split the dataset into groups, and each group contains exactly the same percentage of labels. We used stratified sampling since our dataset is imbalanced. We split the data into five folds for cross-validation.

We have prepared the data for 3-class, 5-class, and 7-class label wise datasets. In 3-class, we have just homophobic, transphobic, and non-anti-LGBT+ content labels. We choose this overall view to see how the system performs for this class. In 5-class, we have taken the homophobic, transphobic, counter speech, hope speech, and none of the above classes. For more fine-grained analysis, we took 7-class where homophobic-derogation, homophobic-threatening, transphobic-derogation, transphobic-threatening, counter speech, hope speech and none of the above. Classwise distribution of the 3-class, 5-class, and 7-class dataset is shown in Table 4, Table 5, and Table 6.

[v]https://en.wikipedia.org/wiki/Arjuna
[w]https://en.wikipedia.org/wiki/Iravan

**Table 7.** Results for English 3 class dataset

| Classifier | Feature | Acc | Pmac | Rmac | F1mac | Pw | Rw | F1w |
|---|---|---|---|---|---|---|---|---|
| LR | TF-IDF (tri-gram) | 0.908 | 0.392 | 0.388 | 0.388 | 0.910 | 0.908 | 0.908 |
| LR | countvec (tri-gram) | 0.916 | 0.388 | 0.378 | 0.382 | 0.908 | 0.916 | 0.912 |
| LR | fastText | 0.638 | 0.366 | 0.480 | 0.328 | 0.924 | 0.638 | 0.734 |
| LR | BERT | 0.904 | 0.442 | 0.504 | 0.466 | 0.926 | 0.904 | 0.914 |
| NB | TF-IDF (tri-gram) | 0.940 | 0.496 | 0.336 | 0.332 | 0.920 | 0.940 | 0.920 |
| NB | countvec (tri-gram) | 0.940 | 0.548 | 0.350 | 0.352 | 0.924 | 0.940 | 0.918 |
| NB | fastText | 0.738 | 0.360 | 0.424 | 0.344 | 0.910 | 0.738 | 0.804 |
| NB | BERT | 0.756 | 0.394 | 0.580 | 0.392 | 0.938 | 0.756 | 0.820 |
| RF | TF-IDF (tri-gram) | 0.812 | 0.426 | 0.362 | 0.354 | 0.908 | 0.812 | 0.832 |
| RF | countvec (tri-gram) | 0.684 | 0.386 | 0.330 | 0.306 | 0.900 | 0.684 | 0.736 |
| RF | fastText | 0.940 | 0.444 | 0.342 | 0.336 | 0.908 | 0.940 | 0.914 |
| RF | BERT | 0.944 | 0.534 | 0.424 | 0.442 | 0.928 | 0.940 | 0.926 |
| SVM | TF-IDF (tri-gram) | 0.934 | 0.422 | 0.370 | 0.380 | 0.910 | 0.934 | 0.920 |
| SVM | countvec (tri-gram) | 0.932 | 0.410 | 0.358 | 0.366 | 0.906 | 0.932 | 0.916 |
| SVM | fastText | 0.940 | 0.310 | 0.330 | 0.320 | 0.890 | 0.940 | 0.910 |
| SVM | BERT | 0.910 | 0.450 | 0.480 | 0.460 | 0.922 | 0.910 | 0.916 |
| DT | TF-IDF (tri-gram) | 0.940 | 0.344 | 0.332 | 0.322 | 0.894 | 0.940 | 0.910 |
| DT | countvec (tri-gram) | 0.940 | 0.412 | 0.334 | 0.326 | 0.904 | 0.940 | 0.912 |
| DT | fastText | 0.940 | 0.310 | 0.330 | 0.320 | 0.890 | 0.940 | 0.910 |
| DT | BERT | 0.934 | 0.400 | 0.374 | 0.380 | 0.910 | 0.934 | 0.918 |
| BiLSTM | - | 0.940 | 0.310 | 0.330 | 0.320 | 0.890 | 0.940 | 0.910 |
| MBERT | - | 0.060 | 0.020 | 0.333 | 0.040 | 0.00 | 0.060 | 0.010 |

### *7.2 Model Building*

Several baseline models are built by using different sets of features and learning algorithms. For both monolingual and code-mixed datasets, the machine learning models with varying embeddings, namely TF-IDF, count vectorizer, BERT (Devlin et al. 2019) embeddings and fastText (Bojanowski et al. 2017) embeddings are used. Classifiers, namely logistic regression, naive Bayes, random forest, support vector machines and decision trees (Pedregosa et al. 2011) are used to build the baseline models with the above embeddings. The models are grouped into three categories, namely

(1) Machine learning models with linguistic features
(2) Machine learning models with word embeddings
(3) Deep learning models

**Table 8.** Results for English 5 class dataset

| Classifier | Feature | Acc | Pmac | Rmac | F1mac | Pw | Rw | F1w |
|---|---|---|---|---|---|---|---|---|
| LR | TF-IDF (tri-gram) | 0.652 | 0.316 | 0.300 | 0.304 | 0.626 | 0.652 | 0.634 |
| LR | countvec (tri-gram) | 0.660 | 0.318 | 0.286 | 0.294 | 0.622 | 0.660 | 0.636 |
| LR | fastText | 0.335 | 0.278 | 0.348 | 0.238 | 0.648 | 0.335 | 0.395 |
| LR | BERT | 0.584 | 0.350 | 0.410 | 0.368 | 0.672 | 0.584 | 0.616 |
| NB | TF-IDF (tri-gram) | 0.708 | 0.378 | 0.210 | 0.182 | 0.614 | 0.708 | 0.592 |
| NB | countvec (tri-gram) | 0.710 | 0.434 | 0.234 | 0.232 | 0.644 | 0.710 | 0.620 |
| NB | fastText | 0.540 | 0.298 | 0.298 | 0.275 | 0.638 | 0.540 | 0.575 |
| NB | BERT | 0.400 | 0.298 | 0.414 | 0.282 | 0.643 | 0.400 | 0.442 |
| RF | TF-IDF (tri-gram) | 0.680 | 0.330 | 0.272 | 0.280 | 0.618 | 0.680 | 0.632 |
| RF | countvec (tri-gram) | 0.674 | 0.318 | 0.266 | 0.278 | 0.610 | 0.675 | 0.632 |
| RF | fastText | 0.688 | 0.313 | 0.220 | 0.210 | 0.583 | 0.688 | 0.600 |
| RF | BERT | 0.698 | 0.378 | 0.262 | 0.280 | 0.628 | 0.698 | 0.636 |
| SVM | TF-IDF (tri-gram) | 0.684 | 0.332 | 0.260 | 0.274 | 0.618 | 0.684 | 0.634 |
| SVM | countvec (tri-gram) | 0.692 | 0.324 | 0.250 | 0.262 | 0.616 | 0.692 | 0.630 |
| SVM | fastText | 0.705 | 0.268 | 0.200 | 0.170 | 0.588 | 0.705 | 0.585 |
| SVM | BERT | 0.612 | 0.350 | 0.356 | 0.350 | 0.632 | 0.612 | 0.620 |
| DT | TF-IDF (tri-gram) | 0.714 | 0.348 | 0.222 | 0.210 | 0.634 | 0.714 | 0.616 |
| DT | countvec (tri-gram) | 0.712 | 0.306 | 0.222 | 0.208 | 0.610 | 0.712 | 0.612 |
| DT | fastText | 0.700 | 0.193 | 0.203 | 0.175 | 0.530 | 0.700 | 0.588 |
| DT | BERT | 0.688 | 0.348 | 0.248 | 0.250 | 0.620 | 0.688 | 0.614 |
| BiLSTM | - | 0.100 | 0.020 | 0.200 | 0.040 | 0.010 | 0.100 | 0.020 |
| MBERT | - | 0.100 | 0.020 | 0.200 | 0.040 | 0.010 | 0.100 | 0.020 |

In the first category, for the linguistic features, n-grams with n are equal to 3 are used, and the text is vectorized using TF-IDF and count vectorizer. In the second category, the text are vectorized using a word embedding features from BERT and FastText. In both these categories, machine learning approaches, namely logistic regression, naive Bayes, random forest, support vector machines and decision trees are used to build the models. Finally, in the last category, the deep learning models, namely Bi-LSTM and MBERT are used.

### *7.2.1 Machine learning with linguistic features*

**Feature Extraction:**

Feature selection is a method that decreases the number of features in the input. To do so, we used several feature extraction techniques, linguistic features such as TF-IDF and count vectorizer in this research.

- **TF-IDF:** TF-IDF is a technique, which generates a numerical weighting of words that represents the importance of each word in a corpus document. This is generally used in the domains such as text mining and information retrieval. We performed TF-IDF with tri-grams. This will extract the word vectors from the corpus by the combination of three words.

**Table 9.** Results for English 7 class dataset

| Classifier | Feature | Acc | Pmac | Rmac | F1mac | Pw | Rw | F1w |
|---|---|---|---|---|---|---|---|---|
| LR | TF-IDF (tri-gram) | 0.648 | 0.240 | 0.234 | 0.234 | 0.626 | 0.648 | 0.636 |
| LR | countvec (tri-gram) | 0.670 | 0.262 | 0.220 | 0.222 | 0.626 | 0.670 | 0.636 |
| LR | fastText | 0.308 | 0.204 | 0.298 | 0.172 | 0.640 | 0.308 | 0.364 |
| LR | BERT | 0.582 | 0.298 | 0.346 | 0.302 | 0.676 | 0.582 | 0.614 |
| NB | TF-IDF (tri-gram) | 0.710 | 0.286 | 0.166 | 0.148 | 0.610 | 0.710 | 0.598 |
| NB | countvec (tri-gram) | 0.710 | 0.372 | 0.198 | 0.204 | 0.642 | 0.710 | 0.626 |
| NB | fastText | 0.516 | 0.210 | 0.228 | 0.194 | 0.634 | 0.516 | 0.558 |
| NB | BERT | 0.384 | 0.220 | 0.352 | 0.210 | 0.646 | 0.384 | 0.430 |
| RF | TF-IDF (tri-gram) | 0.480 | 0.264 | 0.208 | 0.196 | 0.600 | 0.480 | 0.490 |
| RF | countvec (tri-gram) | 0.388 | 0.264 | 0.154 | 0.178 | 0.604 | 0.388 | 0.424 |
| RF | fastText | 0.690 | 0.256 | 0.170 | 0.164 | 0.446 | 0.482 | 0.426 |
| RF | BERT | 0.704 | 0.322 | 0.234 | 0.246 | 0.636 | 0.688 | 0.640 |
| SVM | TF-IDF (tri-gram) | 0.584 | 0.264 | 0.216 | 0.210 | 0.604 | 0.584 | 0.560 |
| SVM | countvec (tri-gram) | 0.698 | 0.282 | 0.216 | 0.220 | 0.618 | 0.698 | 0.632 |
| SVM | fastText | 0.704 | 0.204 | 0.160 | 0.134 | 0.584 | 0.704 | 0.586 |
| SVM | BERT | 0.612 | 0.288 | 0.288 | 0.278 | 0.558 | 0.550 | 0.552 |
| DT | TF-IDF (tri-gram) | 0.712 | 0.264 | 0.176 | 0.162 | 0.624 | 0.712 | 0.612 |
| DT | countvec (tri-gram) | 0.712 | 0.240 | 0.174 | 0.164 | 0.612 | 0.712 | 0.612 |
| DT | fastText | 0.700 | 0.148 | 0.158 | 0.134 | 0.526 | 0.700 | 0.582 |
| DT | BERT | 0.690 | 0.258 | 0.200 | 0.202 | 0.606 | 0.690 | 0.618 |
| BiLSTM | - | 0.700 | 0.112 | 0.158 | 0.132 | 0.500 | 0.700 | 0.580 |
| MBERT | - | 0.100 | 0.010 | 0.140 | 0.030 | 0.010 | 0.100 | 0.020 |

- **Count vectorizer:** Count vectorizer is one of the commonly used feature extraction approaches. This approach converts a text into a vector depending on the frequency with which each word appears throughout the text. Here we performed a count vectorizer with a tri-gram. This will extract the vector with the frequency of a combination of three words. This is done with the help of the sklearn package.

**Classification:**
The vectors are classified using machine learning classifiers like logistic regression, naive Bayes, random forest, support vector machines and decision trees, and the performance is measured.

- **Logistic Regression**
  The classifier makes use of linear combinations of input to predict the output. The probability of the particular class is predicted using the logistic function. In logistic regression, the output depends on the input along with the corresponding system. The estimation of the probability of output is a memory-based method, in which the probability of output $p_o$, the probability of the corresponding input $p_i$, and the probability of system $p_s$. Thus, the classifier predicts the categorical output $p_o$ with $P(p_o/p_i, p_s)$.

**Table 10.** Results for Tamil 3 class dataset

| Classifier | Feature | Acc | Pmac | Rmac | F1mac | Pw | Rw | F1w |
|---|---|---|---|---|---|---|---|---|
| LR | TF-IDF (tri-gram) | 0.836 | 0.690 | 0.598 | 0.632 | 0.824 | 0.836 | 0.824 |
| LR | countvec (tri-gram) | 0.846 | 0.722 | 0.596 | 0.642 | 0.834 | 0.846 | 0.832 |
| LR | fastText | 0.610 | 0.504 | 0.648 | 0.502 | 0.804 | 0.610 | 0.654 |
| LR | BERT | 0.706 | 0.560 | 0.722 | 0.590 | 0.812 | 0.706 | 0.736 |
| NB | TF-IDF (tri-gram) | 0.798 | 0.618 | 0.392 | 0.396 | 0.782 | 0.798 | 0.734 |
| NB | countvec (tri-gram) | 0.824 | 0.790 | 0.476 | 0.522 | 0.820 | 0.824 | 0.786 |
| NB | fastText | 0.720 | 0.544 | 0.648 | 0.568 | 0.798 | 0.720 | 0.748 |
| NB | BERT | 0.532 | 0.466 | 0.466 | 0.408 | 0.718 | 0.532 | 0.574 |
| RF | TF-IDF (tri-gram) | 0.468 | 0.628 | 0.610 | 0.466 | 0.880 | 0.468 | 0.576 |
| RF | countvec (tri-gram) | 0.462 | 0.626 | 0.598 | 0.460 | 0.876 | 0.462 | 0.568 |
| RF | fastText | 0.920 | 0.930 | 0.746 | 0.808 | 0.920 | 0.920 | 0.912 |
| RF | BERT | 0.882 | 0.796 | 0.728 | 0.752 | 0.882 | 0.882 | 0.880 |
| SVM | TF-IDF (tri-gram) | 0.864 | 0.818 | 0.538 | 0.588 | 0.848 | 0.848 | 0.814 |
| SVM | countvec (tri-gram) | 0.855 | 0.815 | 0.575 | 0.638 | 0.853 | 0.855 | 0.835 |
| SVM | fastText | 0.808 | 0.752 | 0.456 | 0.498 | 0.796 | 0.808 | 0.764 |
| SVM | BERT | 0.890 | 0.816 | 0.788 | 0.800 | 0.888 | 0.890 | 0.888 |
| DT | TF-IDF (tri-gram) | 0.810 | 0.652 | 0.418 | 0.420 | 0.772 | 0.792 | 0.732 |
| DT | countvec (tri-gram) | 0.780 | 0.554 | 0.368 | 0.358 | 0.734 | 0.780 | 0.708 |
| DT | fastText | 0.772 | 0.588 | 0.452 | 0.469 | 0.740 | 0.772 | 0.743 |
| DT | BERT | 0.786 | 0.662 | 0.474 | 0.470 | 0.770 | 0.752 | 0.724 |
| BiLSTM | - | 0.890 | 0.300 | 0.330 | 0.310 | 0.800 | 0.890 | 0.840 |
| MBERT | - | 0.168 | 0.328 | 0.434 | 0.282 | 0.252 | 0.168 | 0.142 |

For all the nine data settings, the models were trained using different feature sets with default parameter settings such as C=1.0 and solver='lbfgs'. The number of iterations used to train the models is 100. The class labels are predicted for our set of features with nine settings of data using the models.

- **Naive Bayes:**
  The probabilistic machine learning classifier is based on Acc Theorem. The Bayes theorem states that the probability of a happening process H can be found assuming that the process E has occurred, where E is the evidence and H is the hypothesis with some naive assumptions. The multinomial naive Bayes classifier was trained by learning the prior probabilities from the training set with Alpha value 0.5.
- **Support Vector Machine:**
  The linear classifier makes use of a hyperplane or line to classify the data points. The hyperplane is decided by considering the extreme points, known as support vectors. For the data set D, we need to compute $\gamma = |w \cdot x + b|$ for all training instances to the value G, where G is the smallest $\gamma$. If we have k hyperplanes, each will have a $G_i$ value, and the hyperplane with the largest $G_i$ value will be selected. The models were build using SVM with the parameter values C=0.1, gamma=1 and kernal='linear'.

**Table 11.** Results for Tamil 5 class dataset

| Classifier | Feature | Acc | Pmac | Rmac | F1mac | Pw | Rw | F1w |
|---|---|---|---|---|---|---|---|---|
| LR | TF-IDF (tri-gram) | 0.728 | 0.644 | 0.578 | 0.602 | 0.724 | 0.728 | 0.720 |
| LR | countvec (tri-gram) | 0.750 | 0.690 | 0.584 | 0.620 | 0.746 | 0.750 | 0.738 |
| LR | fastText | 0.428 | 0.416 | 0.550 | 0.404 | 0.672 | 0.428 | 0.448 |
| LR | BERT | 0.574 | 0.528 | 0.630 | 0.514 | 0.688 | 0.574 | 0.598 |
| NB | TF-IDF (tri-gram) | 0.654 | 0.726 | 0.260 | 0.256 | 0.710 | 0.654 | 0.550 |
| NB | countvec (tri-gram) | 0.694 | 0.660 | 0.366 | 0.434 | 0.688 | 0.694 | 0.640 |
| NB | fastText | 0.570 | 0.460 | 0.564 | 0.502 | 0.670 | 0.570 | 0.600 |
| NB | BERT | 0.432 | 0.362 | 0.366 | 0.330 | 0.560 | 0.432 | 0.464 |
| RF | TF-IDF (tri-gram) | 0.494 | 0.668 | 0.556 | 0.520 | 0.788 | 0.494 | 0.558 |
| RF | countvec (tri-gram) | 0.492 | 0.652 | 0.556 | 0.512 | 0.786 | 0.492 | 0.550 |
| RF | fastText | 0.868 | 0.916 | 0.730 | 0.804 | 0.876 | 0.868 | 0.864 |
| RF | BERT | 0.824 | 0.778 | 0.710 | 0.742 | 0.824 | 0.824 | 0.824 |
| SVM | TF-IDF (tri-gram) | 0.782 | 0.826 | 0.574 | 0.642 | 0.792 | 0.782 | 0.760 |
| SVM | countvec (tri-gram) | 0.776 | 0.798 | 0.554 | 0.628 | 0.790 | 0.776 | 0.754 |
| SVM | fastText | 0.678 | 0.740 | 0.336 | 0.368 | 0.702 | 0.678 | 0.608 |
| SVM | BERT | 0.840 | 0.800 | 0.756 | 0.776 | 0.842 | 0.840 | 0.836 |
| DT | TF-IDF (tri-gram) | 0.618 | 0.354 | 0.218 | 0.192 | 0.538 | 0.618 | 0.496 |
| DT | countvec (tri-gram) | 0.622 | 0.406 | 0.220 | 0.194 | 0.554 | 0.622 | 0.496 |
| DT | fastText | 0.614 | 0.394 | 0.294 | 0.294 | 0.548 | 0.614 | 0.558 |
| DT | BERT | 0.618 | 0.524 | 0.276 | 0.282 | 0.582 | 0.616 | 0.538 |
| BiLSTM | - | 0.504 | 0.100 | 0.200 | 0.126 | 0.298 | 0.504 | 0.370 |
| MBERT | - | 0.042 | 0.068 | 0.084 | 0.044 | 0.110 | 0.042 | 0.036 |

- **Decision Tree:**
  A non-parametric model that predicts the target values based on the decision rules concluded from the features of the input vectors. It is a tree-like structure with features being internal nodes, branches being the decision and leaf nodes being the outcomes. The decision tree is based on information gain, entropy and gini index. The decision tree models for all the variations are trained using the parameter values criterion='gini' with minimum sample splits as 2.
- **Random Forest:**
  Random forest is an ensemble model that makes use of a number of decision trees. A vote of the decision of training set trees is aggregated to make the final decision of the test data.

**Table 12.** Results for Tamil 7 class dataset

| Classifier | Feature | Acc | Pmac | Rmac | F1mac | Pw | Rw | F1w |
|---|---|---|---|---|---|---|---|---|
| LR | TF-IDF (tri-gram) | 0.716 | 0.564 | 0.510 | 0.530 | 0.712 | 0.716 | 0.708 |
| LR | countvec (tri-gram) | 0.740 | 0.634 | 0.512 | 0.556 | 0.738 | 0.740 | 0.730 |
| LR | fastText | 0.390 | 0.324 | 0.504 | 0.324 | 0.664 | 0.390 | 0.412 |
| LR | BERT | 0.558 | 0.426 | 0.656 | 0.482 | 0.690 | 0.558 | 0.582 |
| NB | TF-IDF (tri-gram) | 0.648 | 0.550 | 0.188 | 0.190 | 0.690 | 0.648 | 0.544 |
| NB | countvec (tri-gram) | 0.692 | 0.550 | 0.286 | 0.328 | 0.674 | 0.692 | 0.636 |
| NB | fastText | 0.562 | 0.406 | 0.504 | 0.422 | 0.660 | 0.562 | 0.590 |
| NB | BERT | 0.364 | 0.254 | 0.284 | 0.236 | 0.548 | 0.364 | 0.404 |
| RF | TF-IDF (tri-gram) | 0.562 | 0.752 | 0.498 | 0.526 | 0.816 | 0.562 | 0.622 |
| RF | countvec (tri-gram) | 0.482 | 0.706 | 0.500 | 0.500 | 0.806 | 0.482 | 0.562 |
| RF | fastText | 0.850 | 0.936 | 0.704 | 0.788 | 0.870 | 0.862 | 0.856 |
| RF | BERT | 0.824 | 0.738 | 0.676 | 0.688 | 0.836 | 0.824 | 0.822 |
| SVM | TF-IDF (tri-gram) | 0.784 | 0.840 | 0.508 | 0.596 | 0.798 | 0.784 | 0.762 |
| SVM | countvec (tri-gram) | 0.778 | 0.830 | 0.496 | 0.588 | 0.798 | 0.778 | 0.756 |
| SVM | fastText | 0.670 | 0.684 | 0.274 | 0.312 | 0.690 | 0.670 | 0.596 |
| SVM | BERT | 0.850 | 0.808 | 0.746 | 0.770 | 0.848 | 0.850 | 0.848 |
| DT | TF-IDF (tri-gram) | 0.618 | 0.256 | 0.160 | 0.138 | 0.530 | 0.618 | 0.494 |
| DT | countvec (tri-gram) | 0.622 | 0.262 | 0.160 | 0.138 | 0.530 | 0.622 | 0.496 |
| DT | fastText | 0.626 | 0.308 | 0.204 | 0.196 | 0.510 | 0.626 | 0.532 |
| DT | BERT | 0.632 | 0.470 | 0.220 | 0.234 | 0.582 | 0.632 | 0.548 |
| BiLSTM | - | 0.504 | 0.074 | 0.140 | 0.092 | 0.298 | 0.504 | 0.370 |
| MBERT | - | 0.624 | 0.244 | 0.250 | 0.238 | 0.608 | 0.624 | 0.610 |

Like decision tree models, random forest models are also trained using the parameter criterion='gini' with minimum sample splits as 2. In addition, the number of trees in the forest was set to 100.

#### *7.2.2 Machine learning with word embeddings*

**Feature Extraction:**

Word embedding features like BERT embeddings and fastText embeddings are extracted and then classified.

- **BERT:**
  BERT embeddings (Devlin et al. 2019) are the word embeddings that produce vectors that depend on the context of the sentence along with the word of the sentence. BERT is a universal language model that generates a contextualized word embedding at the sub-word level. BERT captures short and long span contextual dependency in the input text via bidirectional self-attention transformers, in contrast to static non-contextualized word embedding. In the BERT embeddings, the sentence is tokenized initially, and [CLS] token is concatenated at the beginning and [SEP] token ending of the sentence. The token embeddings are then generated for each token with a size 768.

**Table 13.** Results for Tamil-English 3 class dataset

| Classifier | Feature | Acc | Pmac | Rmac | F1mac | Pw | Rw | F1w |
|---|---|---|---|---|---|---|---|---|
| LR | TF-IDF (tri-gram) | 0.890 | 0.522 | 0.340 | 0.328 | 0.826 | 0.890 | 0.840 |
| LR | countvec (tri-gram) | 0.890 | 0.508 | 0.340 | 0.326 | 0.822 | 0.890 | 0.840 |
| LR | fastText | 0.584 | 0.396 | 0.530 | 0.370 | 0.860 | 0.584 | 0.672 |
| LR | BERT | 0.812 | 0.490 | 0.604 | 0.524 | 0.876 | 0.812 | 0.838 |
| NB | TF-IDF (tri-gram) | 0.890 | 0.300 | 0.330 | 0.310 | 0.800 | 0.890 | 0.840 |
| NB | countvec (tri-gram) | 0.890 | 0.366 | 0.332 | 0.312 | 0.814 | 0.890 | 0.840 |
| NB | fastText | 0.676 | 0.408 | 0.460 | 0.382 | 0.848 | 0.676 | 0.738 |
| NB | BERT | 0.562 | 0.410 | 0.594 | 0.374 | 0.880 | 0.562 | 0.662 |
| RF | TF-IDF (tri-gram) | 0.428 | 0.478 | 0.356 | 0.184 | 0.864 | 0.428 | 0.434 |
| RF | countvec (tri-gram) | 0.272 | 0.420 | 0.360 | 0.132 | 0.878 | 0.272 | 0.296 |
| RF | fastText | 0.890 | 0.396 | 0.344 | 0.332 | 0.820 | 0.890 | 0.846 |
| RF | BERT | 0.890 | 0.610 | 0.366 | 0.380 | 0.846 | 0.890 | 0.852 |
| SVM | TF-IDF (tri-gram) | 0.890 | 0.530 | 0.338 | 0.324 | 0.826 | 0.890 | 0.840 |
| SVM | countvec (tri-gram) | 0.890 | 0.530 | 0.338 | 0.324 | 0.826 | 0.890 | 0.840 |
| SVM | fastText | 0.890 | 0.398 | 0.338 | 0.326 | 0.808 | 0.890 | 0.842 |
| SVM | BERT | 0.836 | 0.496 | 0.520 | 0.504 | 0.854 | 0.836 | 0.844 |
| DT | TF-IDF (tri-gram) | 0.890 | 0.564 | 0.338 | 0.324 | 0.832 | 0.890 | 0.840 |
| DT | countvec (tri-gram) | 0.890 | 0.498 | 0.336 | 0.322 | 0.818 | 0.890 | 0.840 |
| DT | fastText | 0.890 | 0.332 | 0.332 | 0.314 | 0.802 | 0.890 | 0.840 |
| DT | BERT | 0.888 | 0.620 | 0.380 | 0.396 | 0.850 | 0.888 | 0.850 |
| BiLSTM | - | 0.770 | 0.260 | 0.330 | 0.290 | 0.590 | 0.770 | 0.670 |
| MBERT | - | 0.030 | 0.010 | 0.330 | 0.020 | 0.000 | 0.030 | 0.000 |

- **fastText:**
  fastText (Grave et al. (2018)) is a pre-trained vector model that was trained on Common Crawl and Wikipedia data for 157 languages with Continuous Bag-Of-Words (CBOW) position weights in 300 dimensions, which learns sub-word information and allows users to construct representations for uncommon or out-of-vocabulary terms. For a set of n documents, the model will minimize the log-likelihood over the classes by decaying the learning rate with normalized bag-of-words of $n^{th}$ document $i_n$ and label $o_n$ along with the weight matrices X and Y.

$$-\frac{1}{N}\sum_{n=1}^{N} i_n \, log(f(XYo_n)) \tag{5}$$

**Table 14.** Results for Tamil-English 5 class dataset

| Classifier | Feature | Acc | Pmac | Rmac | F1mac | Pw | Rw | F1w |
|---|---|---|---|---|---|---|---|---|
| LR | TF-IDF (tri-gram) | 0.794 | 0.532 | 0.238 | 0.246 | 0.722 | 0.794 | 0.718 |
| LR | countvec (tri-gram) | 0.794 | 0.512 | 0.240 | 0.242 | 0.716 | 0.794 | 0.718 |
| LR | fastText | 0.442 | 0.300 | 0.454 | 0.294 | 0.758 | 0.442 | 0.520 |
| LR | BERT | 0.672 | 0.386 | 0.508 | 0.420 | 0.776 | 0.672 | 0.712 |
| NB | TF-IDF (tri-gram) | 0.790 | 0.160 | 0.200 | 0.180 | 0.630 | 0.790 | 0.700 |
| NB | countvec (tri-gram) | 0.790 | 0.340 | 0.202 | 0.182 | 0.680 | 0.790 | 0.702 |
| NB | fastText | 0.608 | 0.310 | 0.370 | 0.298 | 0.738 | 0.608 | 0.654 |
| NB | BERT | 0.428 | 0.308 | 0.476 | 0.286 | 0.784 | 0.428 | 0.510 |
| RF | TF-IDF (tri-gram) | 0.384 | 0.518 | 0.244 | 0.162 | 0.778 | 0.384 | 0.374 |
| RF | countvec (tri-gram) | 0.520 | 0.522 | 0.244 | 0.188 | 0.748 | 0.520 | 0.486 |
| RF | fastText | 0.792 | 0.376 | 0.230 | 0.228 | 0.694 | 0.792 | 0.716 |
| RF | BERT | 0.802 | 0.592 | 0.276 | 0.310 | 0.752 | 0.802 | 0.742 |
| SVM | TF-IDF (tri-gram) | 0.800 | 0.554 | 0.234 | 0.238 | 0.724 | 0.800 | 0.718 |
| SVM | countvec (tri-gram) | 0.798 | 0.550 | 0.228 | 0.232 | 0.724 | 0.798 | 0.718 |
| SVM | fastText | 0.794 | 0.416 | 0.214 | 0.202 | 0.688 | 0.794 | 0.708 |
| SVM | BERT | 0.738 | 0.408 | 0.412 | 0.416 | 0.746 | 0.738 | 0.742 |
| DT | TF-IDF (tri-gram) | 0.798 | 0.384 | 0.218 | 0.216 | 0.692 | 0.798 | 0.712 |
| DT | countvec (tri-gram) | 0.798 | 0.344 | 0.214 | 0.210 | 0.680 | 0.798 | 0.710 |
| DT | fastText | 0.790 | 0.160 | 0.200 | 0.180 | 0.630 | 0.772 | 0.700 |
| DT | BERT | 0.788 | 0.430 | 0.254 | 0.268 | 0.702 | 0.788 | 0.722 |
| BiLSTM | - | 0.050 | 0.010 | 0.200 | 0.020 | 0.000 | 0.050 | 0.000 |
| MBERT | - | 0.044 | 0.064 | 0.242 | 0.074 | 0.012 | 0.044 | 0.016 |

**Classification:**
The vectors are classified using traditional classifiers as in the first category. A total of 180 models were trained with a combination of five classifiers using four sets of features for the nine data settings. To assess the consistency of the performance, we have also performed 5-fold cross-validation on training data.

#### *7.2.3 Deep learning models*

- **Bi-LSTM:** The deep learning model with a Bi-directional LSTM layer was built. The model had an embedding layer at the top where the input vectors are vectorized using BERT embeddings followed by the Bi-LSTM layer, flatten layer and two dense layers. It was implemented using Keras layers (Chollet et al. 2015). Finally, the probability distribution over the classification classes is produced using a linear, fully connected layer with the Softmax activation function, and the class with the highest probability is chosen as the final label. We trained all our machine learning and deep learning models on the Google Colab Pro [x].

[x]https://colab.research.google.com/

**Table 15.** Results for Tamil-English 7 class dataset

| Classifier | Feature | Acc | Pmac | Rmac | F1mac | Pw | Rw | F1w |
|---|---|---|---|---|---|---|---|---|
| LR | TF-IDF (tri-gram) | 0.794 | 0.532 | 0.238 | 0.246 | 0.722 | 0.794 | 0.718 |
| LR | countvec (tri-gram) | 0.794 | 0.512 | 0.240 | 0.242 | 0.716 | 0.794 | 0.718 |
| LR | fastText | 0.442 | 0.300 | 0.454 | 0.294 | 0.758 | 0.442 | 0.520 |
| LR | BERT | 0.672 | 0.386 | 0.508 | 0.420 | 0.776 | 0.672 | 0.712 |
| NB | TF-IDF (tri-gram) | 0.790 | 0.160 | 0.200 | 0.180 | 0.630 | 0.790 | 0.700 |
| NB | countvec (tri-gram) | 0.790 | 0.340 | 0.202 | 0.182 | 0.680 | 0.790 | 0.702 |
| NB | fastText | 0.608 | 6.052 | 0.370 | 0.298 | 0.738 | 0.608 | 0.654 |
| NB | BERT | 0.428 | 0.308 | 0.476 | 0.286 | 0.784 | 0.428 | 0.510 |
| RF | TF-IDF (tri-gram) | 0.384 | 0.518 | 0.244 | 0.162 | 0.778 | 0.384 | 0.374 |
| RF | countvec (tri-gram) | 0.520 | 0.522 | 0.244 | 0.188 | 0.748 | 0.520 | 0.486 |
| RF | fastText | 0.792 | 0.376 | 0.230 | 0.228 | 0.694 | 0.792 | 0.716 |
| RF | BERT | 0.802 | 0.592 | 0.276 | 0.310 | 0.752 | 0.802 | 0.742 |
| SVM | TF-IDF (tri-gram) | 0.800 | 0.554 | 0.234 | 0.238 | 0.724 | 0.800 | 0.718 |
| SVM | countvec (tri-gram) | 0.798 | 0.550 | 0.228 | 0.232 | 0.724 | 0.798 | 0.718 |
| SVM | fastText | 0.794 | 0.416 | 0.214 | 0.202 | 0.688 | 0.794 | 0.708 |
| SVM | BERT | 0.738 | 0.408 | 0.412 | 0.416 | 0.746 | 0.738 | 0.742 |
| DT | TF-IDF (tri-gram) | 0.798 | 0.384 | 0.218 | 0.216 | 0.692 | 0.798 | 0.712 |
| DT | countvec (tri-gram) | 0.798 | 0.344 | 0.214 | 0.210 | 0.680 | 0.798 | 0.710 |
| DT | fastText | 0.790 | 0.160 | 0.200 | 0.180 | 0.630 | 0.772 | 0.700 |
| DT | BERT | 0.788 | 0.430 | 0.254 | 0.268 | 0.702 | 0.788 | 0.722 |
| BiLSTM | - | 0.790 | 0.110 | 0.140 | 0.130 | 0.630 | 0.790 | 0.700 |
| MBERT | - | 0.032 | 0.060 | 0.146 | 0.042 | 0.096 | 0.032 | 0.018 |

- **mBERT:** Bidirectional Encoder Representations from Transformers (BERT) makes use of an attention mechanisms to classify the input. It has an encoder that reads the input and a decoder that produces the output. It uses contextual relations between the words for classification since it read the entire sequence as its input instead of reading from a particular direction. For the classification task, a classification layer added to the existing architecture. Multilingual BERT (mBERT) was trained on 104 languages other than English to learn multi-lingual and cross representation better than other representations. Thus, mBERT was used to classify the text.

## 8. Evaluation

### *8.1 Evaluation metrics*

Following the selection of the method and construction of the classifier, the performance of the classification model must be assessed in order to determine the classification model's capacity to classify unknown data into the proper classes. We utilized a variety of ways to assess the classification algorithm's performance, including accuracy, precision, and recall, as well as the F1-score, which are defined as follows:

$$\text{Recall (R)} = \frac{TP}{(TP + FN)} \tag{6}$$

$$\text{Precision (P)} = \frac{TP}{(TP + FP)} \tag{7}$$

$$\text{Accuracy} = \frac{(TP + TN)}{(TP + TN + FP + FN)} \tag{8}$$

$$F1 = \frac{(2 \times \text{Precision} \times \text{Recall})}{(\text{Precision} + \text{Recall})} \tag{9}$$

$$P_{\text{mac}} = \frac{1}{L} \sum_{i=1}^{L} P \text{ of } i \tag{10}$$

$$R_{\text{mac}} = \frac{1}{L} \sum_{i=1}^{L} R \text{ of } i \tag{11}$$

$$F1_{\text{mac}} = \frac{1}{L} \sum_{i=1}^{L} 2 \times \frac{P_{mac} \times R_{\text{mac}}}{P_{mac} + R_{\text{mac}}} \tag{12}$$

$$P_{mic} = \frac{\sum_{i=1}^{L} \text{tp of label } i}{\sum_{i=1}^{L} (\text{tp of label } i + fp \text{ of label } i)} \tag{13}$$

$$R_{mic} = \frac{\sum_{i=1}^{L} tp \text{ of label } i}{\sum_{i=1}^{L} (\text{tp of label } i + fn \text{ of label } i)} \tag{14}$$

$$F1_{mic} = 2 \times \frac{P_{\text{mic}} \times R_{mic}}{P_{\text{mic}} + R_{mic}} \tag{15}$$

$$P_{\text{weighted}} = \sum_{i=1}^{L} (P \text{ of } i \times \text{Weight of } i) \tag{16}$$

$$R_{\text{weighted}} = \sum_{i=1}^{L} (R \text{ of } i \times \text{Weight of } i) \tag{17}$$

$$F1_{\text{weighted}} = \sum_{i=1}^{L} (F1 \text{ of } i \times \text{Weight of } i) \tag{18}$$

where TP, TN, FP, and FN refer to True Positive, True Negative, False Positive, and False Negative respectively. We present the total F1-macro obtained with various features across all classification levels since we are working with unbalanced data in the 2-class, 3-class, and 7-class datasets. Micro-F is comparable to accuracy in classification jobs when each test case is guaranteed to be given one and only one class. Micro-F averages all instances in the test set, whereas macro-F averages all instances per class first and then all class results. As a result, every class receives an identical weight in macro-F, but the weight in micro-F is proportional to the number of cases in the test data. The latter indicates that minority classes have the same effect on accuracy as majority classes in macro-F, but mistakes in minority class instances have less impact on accuracy in micro-F due to the small number of such cases. So we gave we calculated macro average and weighted average scores.

**Table 16.** Statistical Test Results for English 3 class dataset

| Models | Chi-Test $\chi^2_{critical}$ (df-4, alpha-0.05) = 0.949 $\chi^2_{statistic}$ of | | | | T - Test | | | |
|---|---|---|---|---|---|---|---|---|
| | Accuracy | Precision | Recall | F1-score | Accuracy | Precision | Recall | F1-score |
| Tf_LR | 1.00E+00 | 9.94E-01 | 1.00E+00 | 1.00E+00 | 4.22E-04 | 4.51E-03 | 4.76E-02 | 1.56E-01 |
| Tf_NB | 1.00E+00 | 9.83E-01 | 1.00E+00 | 9.99E-01 | 8.89E-02 | 4.83E-01 | 2.68E-03 | 1.94E-03 |
| Tf_RF | 8.90E-01 | 9.98E-01 | 1.00E+00 | 9.99E-01 | 1.66E-01 | 1.24E-02 | 5.01E-02 | 7.09E-02 |
| Tf_SVM | 1.00E+00 | 9.97E-01 | 1.00E+00 | 1.00E+00 | 4.45E-02 | 3.40E-02 | 1.99E-01 | 1.42E-01 |
| Tf_DT | 1.00E+00 | 9.67E-01 | 1.00E+00 | 9.99E-01 | 8.89E-02 | 1.38E-02 | 1.25E-03 | 7.58E-04 |
| Count_LR | 1.00E+00 | 9.95E-01 | 1.00E+00 | 1.00E+00 | 8.53E-04 | 2.97E-03 | 4.09E-01 | 4.43E-02 |
| Count_NB | 1.00E+00 | 9.99E-01 | 1.00E+00 | 1.00E+00 | 8.89E-02 | 2.53E-01 | 3.46E-03 | 2.47E-03 |
| Count_RF | 6.62E-01 | 9.95E-01 | 9.99E-01 | 9.89E-01 | 7.58E-02 | 2.03E-03 | 4.45E-02 | 2.63E-02 |
| Count_SVM | 1.00E+00 | 9.97E-01 | 1.00E+00 | 1.00E+00 | 1.63E-02 | 1.41E-02 | 3.14E-02 | 2.98E-02 |
| Count_DT | 1.00E+00 | 9.77E-01 | 1.00E+00 | 9.99E-01 | 8.89E-02 | 1.37E-01 | 1.24E-03 | 7.74E-04 |
| Bert_LR | 1.00E+00 | 9.98E-01 | 9.96E-01 | 9.99E-01 | 4.32E-04 | 9.52E-02 | 1.01E-02 | 5.50E-02 |
| Bert_NB | 9.94E-01 | 9.95E-01 | 9.85E-01 | 1.00E+00 | 2.76E-06 | 5.92E-03 | 2.24E-03 | 1.94E-01 |
| Bert_SVM | 1.00E+00 | 9.97E-01 | 9.98E-01 | 9.99E-01 | 5.23E-03 | 1.67E-01 | 2.06E-02 | 9.00E-02 |
| Bert_DT | 1.00E+00 | 9.99E-01 | 1.00E+00 | 1.00E+00 | 9.45E-02 | 1.18E-02 | 1.25E-01 | 3.75E-01 |
| Fast_LR | 9.47E-01 | 9.89E-01 | 9.98E-01 | 9.99E-01 | 2.31E-07 | 2.50E-03 | 1.55E-02 | 8.77E-04 |
| Fast_NB | 9.90E-01 | 9.86E-01 | 1.00E+00 | 1.00E+00 | 7.18E-06 | 1.98E-03 | 2.18E-02 | 6.06E-03 |
| Fast_RF | 1.00E+00 | 9.98E-01 | 1.00E+00 | 1.00E+00 | 8.89E-02 | 5.38E-02 | 3.03E-03 | 1.64E-03 |
| Fast_SVM | 1.00E+00 | 9.57E-01 | 1.00E+00 | 9.99E-01 | 8.89E-02 | 6.28E-04 | 1.09E-03 | 5.56E-04 |
| Fast_DT | 1.00E+00 | 9.57E-01 | 1.00E+00 | 9.99E-01 | 8.89E-02 | 6.28E-04 | 1.09E-03 | 5.56E-04 |
| Bi-Lstm | 1.00E+00 | 9.57E-01 | 1.00E+00 | 9.99E-01 | 8.89E-02 | 6.28E-04 | 1.09E-03 | 5.56E-04 |
| MBERT | 2.42E-13 | 1.93E-03 | 7.67E-02 | 2.11E-15 | 1.77E-10 | 2.45E-05 | 4.84E-07 | 7.86E-07 |

## 9. Results

In this section, we show the results of the baseline models for our dataset. In addition, we report the results in accuracy, macro average precision, macro average recall, macro average F1 score, weighted average precision, weighted average recall, and weighted average F1 score. Table 7, Table 10, Table 13, Table 8, Table 11, Table 14, Table 9, Table 12, and Table 15 presents the classification performance of the machine learning algorithms and deep learning models combined with different features for the 3-class dataset. Accuracy ranges between 0.63 to 0.94 for 3-class English; it drops down to 0.33 to 0.71 when the data is fine-grained to 5-class labels. For 7-class it is more or less close to 5-class labels between 0.30 to 0.71. Over all macro average score is below 0.4 for all precision, recall, and F1 score for all 3-class, 5-class and 7-class settings. As macro

**Table 17.** Statistical Test Results for English 5 class dataset

| Models | Chi-Test $\chi^2_{critical}$ (df-4, alpha-0.05) = 0.949 $\chi^2_{statistic}$ of | | | | T - Test | | | |
|---|---|---|---|---|---|---|---|---|
| | Accuracy | Precision | Recall | F1-score | Accuracy | Precision | Recall | F1-score |
| Tf_LR | 1.00E+00 | 9.95E-01 | 9.98E-01 | 9.86E-01 | 1.63E-04 | 6.83E-04 | 1.18E-02 | 2.54E-04 |
| Tf_NB | 1.00E+00 | 9.98E-01 | 1.00E+00 | 9.97E-01 | 4.45E-02 | 1.27E-03 | 1.04E-04 | 1.83E-04 |
| Tf_RF | 1.00E+00 | 9.99E-01 | 1.00E+00 | 9.68E-01 | 1.07E-02 | 1.63E-02 | 1.75E-04 | 5.29E-05 |
| Tf_SVM | 1.00E+00 | 9.87E-01 | 9.99E-01 | 8.91E-01 | 5.40E-02 | 8.84E-03 | 3.14E-04 | 3.08E-05 |
| Tf_DT | 1.00E+00 | 8.70E-01 | 9.99E-01 | 9.76E-01 | 8.07E-03 | 3.77E-04 | 4.64E-05 | 7.64E-05 |
| Count_LR | 1.00E+00 | 1.00E+00 | 1.00E+00 | 9.99E-01 | 2.26E-05 | 3.99E-03 | 2.10E-04 | 2.00E-03 |
| Count_NB | 1.00E+00 | 1.00E+00 | 1.00E+00 | 9.87E-01 | 1.94E-03 | 2.03E-02 | 6.14E-05 | 5.20E-04 |
| Count_RF | 1.00E+00 | 9.99E-01 | 1.00E+00 | 9.97E-01 | 4.63E-03 | 1.16E-02 | 2.07E-01 | 1.78E-03 |
| Count_SVM | 1.00E+00 | 9.99E-01 | 1.00E+00 | 9.93E-01 | 1.52E-01 | 3.03E-02 | 1.55E-02 | 3.84E-04 |
| Count_DT | 1.00E+00 | 9.95E-01 | 1.00E+00 | 9.59E-01 | 2.32E-03 | 4.84E-02 | 5.27E-05 | 8.45E-05 |
| Bert_LR | 9.98E-01 | 1.00E+00 | 9.92E-01 | 1.00E+00 | 1.18E-05 | 2.61E-02 | 1.57E-04 | #NUM! |
| Bert_NB | 8.87E-01 | 9.98E-01 | 9.90E-01 | 9.98E-01 | 1.41E-05 | 3.99E-04 | 9.66E-04 | 6.36E-04 |
| Bert_SVM | 9.99E-01 | 1.00E+00 | 9.98E-01 | 1.00E+00 | 1.62E-03 | 8.62E-02 | 8.89E-04 | 2.78E-02 |
| Bert_DT | 1.00E+00 | 1.00E+00 | 1.00E+00 | 9.87E-01 | 1.71E-02 | 1.18E-01 | 2.10E-01 | 3.29E-03 |
| Fast_LR | 7.51E-01 | 9.95E-01 | 9.98E-01 | 9.86E-01 | 1.79E-08 | 6.83E-04 | 1.18E-02 | 2.54E-04 |
| Fast_NB | 9.93E-01 | 9.98E-01 | 1.00E+00 | 9.97E-01 | 4.09E-06 | 1.27E-03 | 1.04E-04 | 1.83E-04 |
| Fast_RF | 1.00E+00 | 9.99E-01 | 1.00E+00 | 9.68E-01 | 4.97E-02 | 1.63E-02 | 1.75E-04 | 5.29E-05 |
| Fast_SVM | 1.00E+00 | 9.87E-01 | 9.99E-01 | 8.91E-01 | 1.52E-01 | 8.84E-03 | 3.14E-04 | 3.08E-05 |
| Fast_DT | 1.00E+00 | 8.70E-01 | 9.99E-01 | 8.90E-01 | 3.11E-01 | 3.77E-04 | 4.64E-05 | 3.34E-05 |
| Bi-Lstm | 1.30E-03 | 1.69E-06 | 9.99E-01 | 5.82E-03 | 4.60E-09 | 4.33E-06 | 1.28E-05 | 4.86E-10 |
| MBERT | 1.30E-03 | 1.69E-06 | 9.99E-01 | 9.28E-03 | 4.60E-09 | 4.33E-06 | 1.28E-05 | 4.59E-06 |

df - degree of freedom

average penalizes when a model does not perform well with minority classes, and our dataset is highly imbalanced, we believe that there is a more open area for future research on finding the perfect model.

For Tamil, the accuracy ranges from 0.61 to 0.92 for 3-class labels, 0.42 to 0.86 for 5-class labels, and 0.39 to 0.85 for 7-class labels. The accuracy does not drop so much as compared to English when the data is more fine-grained. From the tables, we can see that fastText with RF performs best for all three label settings for Tamil in terms of features. The relatively lower classification results for the task at 7-class is a consequence of the increased granularity. We can observe that among the model+feature combination Random Forest with BERT embedding has the highest weighted F1 score for the English and Tamil-English code mixed settings. For the Tamil language, the highest weighted F1 score is from Random Forest with fastText embedding combination. Tamil-English code-mixed settings for all three settings are almost similar to English due to the presence of English words and romanized script. From our experiment results for all three languages and three class label settings, hybrid deep learning and machine learning worked better than pure machine learning or deep learning. We have experimented with only BiLSTM and Multilingual BERT for deep learning settings. In some of the settings Multilingual BERT performed poorer compared to all other classifiers. Across all the nine settings, we observe that an English 7 class setting has a low weighted F1 score due to the high-class imbalance of the

**Table 18.** Statistical Test Results for English 7 class dataset

| Models | Chi-Test $\chi^2_{critical}$ (df-4, alpha-0.05) = 0.949 $\chi^2_{statistic}$ of | | | | T - Test | | | |
|---|---|---|---|---|---|---|---|---|
| | Accuracy | Precision | Recall | F1-score | Accuracy | Precision | Recall | F1-score |
| Tf_LR | 1.00E+00 | 9.99E-01 | 8.78E-01 | 1.00E+00 | 7.55E-05 | 3.60E-02 | 1.86E-02 | 9.68E-02 |
| Tf_NB | 1.00E+00 | 9.98E-01 | 1.67E-04 | 9.94E-01 | 1.04E-01 | 1.98E-01 | 2.36E-02 | 1.14E-03 |
| Tf_RF | 5.88E-01 | 9.99E-01 | 9.99E-01 | 1.00E+00 | 6.38E-02 | 3.58E-02 | 3.94E-01 | 4.33E-02 |
| Tf_SVM | 8.40E-01 | 9.99E-01 | 1.00E+00 | 1.00E+00 | 1.48E-01 | 3.86E-02 | 3.84E-01 | 1.68E-01 |
| Tf_DT | 1.00E+00 | 9.90E-01 | 9.99E-01 | 9.95E-01 | 4.97E-02 | 1.35E-01 | 1.32E-02 | 9.72E-03 |
| Count_LR | 1.00E+00 | 9.99E-01 | 1.00E+00 | 1.00E+00 | 1.28E-02 | 1.72E-01 | 1.26E-01 | 2.86E-01 |
| Count_NB | 1.00E+00 | 9.97E-01 | 1.00E+00 | 9.99E-01 | 1.04E-01 | 2.42E-01 | 1.72E-01 | 1.61E-01 |
| Count_RF | 3.82E-01 | 9.99E-01 | 9.82E-01 | 9.96E-01 | 2.96E-02 | 1.12E-01 | 1.10E-01 | 1.02E-01 |
| Count_SVM | 1.00E+00 | 1.00E+00 | 1.00E+00 | 1.00E+00 | 1.52E-01 | 5.68E-02 | 4.25E-01 | 1.40E-01 |
| Count_DT | 1.00E+00 | 9.96E-01 | 9.99E-01 | 9.96E-01 | 4.97E-02 | 1.01E-02 | 9.28E-03 | 6.26E-03 |
| Bert_LR | 9.98E-01 | 9.98E-01 | 9.88E-01 | 9.97E-01 | 7.18E-05 | 4.84E-01 | 3.51E-03 | 2.86E-02 |
| Bert_NB | 8.47E-01 | 9.96E-01 | 9.86E-01 | 1.00E+00 | 2.05E-05 | 1.95E-02 | 3.01E-03 | 4.37E-01 |
| Bert_SVM | 9.99E-01 | 9.98E-01 | 9.98E-01 | 9.99E-01 | 8.83E-04 | 3.66E-01 | 1.45E-02 | 4.09E-02 |
| Bert_DT | 1.00E+00 | 5.68E-01 | 1.00E+00 | 9.98E-01 | 5.40E-02 | 2.08E-01 | 2.00E-01 | 1.43E-01 |
| Fast_LR | 6.36E-01 | 9.90E-01 | 9.96E-01 | 9.99E-01 | 3.12E-08 | 1.59E-02 | 6.71E-03 | 2.44E-02 |
| Fast_NB | 9.87E-01 | 9.91E-01 | 1.00E+00 | 1.00E+00 | 6.93E-06 | 2.95E-02 | 8.43E-02 | 1.72E-01 |
| Fast_RF | 1.00E+00 | 9.95E-01 | 9.99E-01 | 9.97E-01 | 2.32E-03 | 8.30E-02 | 7.42E-03 | 6.59E-03 |
| Fast_SVM | 1.00E+00 | 9.76E-01 | 9.98E-01 | 9.84E-01 | 5.00E-01 | 2.01E-02 | 4.53E-03 | 2.41E-03 |
| Fast_DT | 1.00E+00 | 8.21E-01 | 9.98E-01 | 9.81E-01 | 8.89E-02 | 6.75E-03 | 5.15E-03 | 3.50E-03 |
| Bi-Lstm | 1.00E+00 | 7.17E-01 | 9.98E-01 | 9.79E-01 | 1.87E-01 | 1.47E-03 | 4.41E-03 | 2.62E-03 |
| MBERT | 1.11E-03 | 1.17E-08 | 9.98E-01 | 4.46E-02 | 5.77E-09 | 3.52E-04 | 3.71E-03 | 1.81E-04 |

df - degree of freedom

dataset. As we can see from the Table 6, the transphobic-threatening class has only two instances for English while Tamil and Tamil-English setting has more than 50. Instances for homophobic-threatening and transphobic-derogation are below 15 for English data, while it is more than 50 for Tamil and Tamil-English settings. This class imbalance also affected the performance of our classifiers.

All feature selection methods and classifiers for both English and Tamil-English settings macro average scores of precision, recall, and F1-score were below 0.5. However, for Tamil dataset, classifiers score better, and some classifiers above 0.8 macro averaged F1-score. The low scores in Tamil-English are due to the problem of code-mixing at different levels. Even-though we trained our systems with multilingual settings, the code-mixing in our dataset is unpredictable, there is no standard spelling when social media users try to write Tamil Roman script, and the dialectal variants influence the morphology changes code-mixed words.

### *9.1 Statistical Test*

To evaluate whether there is a difference between statistically significant algorithms, we conducted a T-test. T-test (One-tail, Type 1 and alpha 0.5) was performed to analyse the best significant model

**Table 19.** Statistical Test Results for Tamil 3 class dataset

| Models | Chi-Test $\chi^2_{critical}$ (df-4, alpha-0.05) = 0.949 $\chi^2_{statistic}$ of | | | | T - Test | | | |
|---|---|---|---|---|---|---|---|---|
| | Accuracy | Precision | Recall | F1-score | Accuracy | Precision | Recall | F1-score |
| Tf_LR | 1.00E+00 | 9.79E-01 | 9.96E-01 | 9.92E-01 | 7.55E-05 | 1.53E-04 | 5.26E-04 | 2.62E-04 |
| Tf_NB | 9.99E-01 | 9.26E-01 | 8.15E-01 | 7.01E-01 | 2.64E-06 | 5.41E-04 | 1.30E-05 | 1.08E-05 |
| Tf_RF | 7.00E-01 | 9.45E-01 | 9.98E-01 | 8.69E-01 | 2.09E-07 | 1.05E-05 | 1.30E-04 | 1.10E-05 |
| Tf_SVM | 1.00E+00 | 1.00E+00 | 9.95E-01 | 9.96E-01 | 7.55E-05 | 8.77E-03 | 2.37E-04 | 3.79E-04 |
| Tf_DT | 9.99E-01 | 8.83E-01 | 7.47E-01 | 5.64E-01 | 2.09E-03 | 1.86E-02 | 5.64E-06 | 4.54E-06 |
| Count_LR | 1.00E+00 | 9.88E-01 | 9.96E-01 | 9.94E-01 | 2.51E-05 | 4.05E-04 | 8.70E-04 | 6.27E-04 |
| Count_NB | 1.00E+00 | 9.97E-01 | 9.48E-01 | 9.39E-01 | 2.34E-05 | 7.83E-03 | 1.82E-05 | 6.92E-06 |
| Count_RF | 6.85E-01 | 9.45E-01 | 9.97E-01 | 8.55E-01 | 7.87E-08 | 6.45E-06 | 6.38E-05 | 9.86E-06 |
| Count_SVM | 1.00E+00 | 9.99E-01 | 9.94E-01 | 9.94E-01 | 4.46E-05 | 3.39E-03 | 3.02E-04 | 5.37E-04 |
| Count_DT | 9.98E-01 | 7.91E-01 | 7.51E-01 | 5.77E-01 | 6.96E-06 | 3.22E-03 | 1.09E-05 | 9.22E-06 |
| Bert_LR | 9.88E-01 | 8.69E-01 | 1.00E+00 | 9.82E-01 | 4.45E-06 | 8.78E-07 | 9.30E-02 | 6.09E-06 |
| Bert_NB | 8.38E-01 | 6.77E-01 | 9.56E-01 | 7.95E-01 | 2.73E-06 | 1.64E-07 | 5.54E-03 | 2.56E-04 |
| Bert_RF | 1.00E+00 | 9.97E-01 | 1.00E+00 | 1.00E+00 | 4.49E-03 | 5.97E-03 | 1.91E-01 | 1.68E-02 |
| Bert_SVM | 1.00E+00 | 9.99E-01 | 1.00E+00 | 1.00E+00 | 6.62E-03 | 1.80E-04 | 9.44E-03 | 3.08E-01 |
| Bert_DT | 9.98E-01 | 9.81E-01 | 8.99E-01 | 8.74E-01 | 3.34E-07 | 4.59E-04 | 8.40E-06 | 7.33E-06 |
| Fast_LR | 9.40E-01 | 7.69E-01 | 9.99E-01 | 9.22E-01 | 1.30E-07 | 5.33E-07 | 1.86E-04 | 4.69E-06 |
| Fast_NB | 9.91E-01 | 8.46E-01 | 9.99E-01 | 9.71E-01 | 7.48E-07 | 3.85E-07 | 1.46E-04 | 3.26E-06 |
| Fast_SVM | 9.99E-01 | 9.92E-01 | 9.23E-01 | 9.09E-01 | 2.16E-05 | 2.13E-03 | 4.27E-05 | 7.88E-05 |
| Fast_DT | 9.98E-01 | 9.00E-01 | 9.20E-01 | 8.77E-01 | 1.22E-06 | 1.22E-04 | 6.11E-05 | 3.71E-05 |
| Bi-Lstm | 9.97E-01 | 6.38E-02 | 6.45E-01 | 3.26E-01 | 5.91E-07 | 1.52E-07 | 3.48E-06 | 1.35E-06 |
| MBERT | 1.92E-03 | 1.62E-01 | 8.46E-01 | 3.06E-01 | 2.73E-08 | 1.90E-04 | 1.13E-03 | 1.96E-06 |

df - degree of freedom

by comparing the best models with other models.

$$T_{test} = \frac{\bar{x}_1 - \bar{x}_2}{\sqrt{s^2\left(\frac{1}{n_1}+\frac{1}{n_2}\right)}} \quad (19)$$

where $\bar{x}_1$ and $\bar{x}_2$ be the means of the two groups namely the best model and the model to be com- pared. $n_1$ and $n_2$ be the number of observations in the group and $s$ be the standard error between these two groups. Before applying t-test, it is necessary to confirm that the groups are normally distributed. So, we have performed chi-square test on the two groups. As the population size is five for the groups, the $\chi^2_{stat}$ values were computed and compared with $\chi^2_{critical}$ values with alpha 0.05 and four degree of freedom. From the experiments, it is observed that all the values are lesser than 0.949 and thus, t-test can be performed. We chose the population as the five-fold validation results in accuracy / precision/ recall / F1-scores. For example, the F1-scores obtained from five folds of the best model and the other models are applied for the t-test. If the t-test score is lesser than 0.05, we can conclude that the improvement is statistically significant. From the experimental results, it is observed that random forest performs better than all the other classifiers. The models with BERT embedding features give better results for English and Tamil-English code mixed languages, and the model with fasttext embedding features gives good results for the Tamil language. Statistical test, namely k-paired t-test, was conducted between the best model and all the

**Table 20.** Statistical Test Results for Tamil 5 class dataset

| Models | Chi-Test $\chi^2_{critical}$ (df-4, alpha-0.05) = 0.949 $\chi^2_{statistic}$ of | | | | T - Test | | | |
|---|---|---|---|---|---|---|---|---|
| | Accuracy | Precision | Recall | F1-score | Accuracy | Precision | Recall | F1-score |
| Tf_LR | 9.98E-01 | 9.65E-01 | 9.95E-01 | 9.88E-01 | 3.74E-05 | 6.41E-06 | 8.36E-04 | 1.85E-04 |
| Tf_NB | 9.86E-01 | 9.87E-01 | 3.73E-01 | 2.27E-01 | 9.63E-07 | 8.24E-03 | 6.64E-06 | 2.41E-06 |
| Tf_RF | 7.72E-01 | 9.78E-01 | 9.91E-01 | 9.46E-01 | 2.12E-03 | 2.41E-06 | 9.07E-04 | 5.48E-05 |
| Tf_SVM | 1.00E+00 | 9.99E-01 | 9.94E-01 | 9.96E-01 | 3.62E-05 | 1.30E-04 | 1.96E-04 | 1.28E-04 |
| Tf_DT | 9.73E-01 | 3.06E-01 | 2.04E-01 | 4.76E-02 | 7.67E-08 | 3.18E-05 | 5.14E-06 | 1.61E-06 |
| Count_LR | 9.99E-01 | 9.85E-01 | 9.96E-01 | 9.92E-01 | 4.40E-05 | 3.45E-05 | 6.96E-04 | 1.41E-04 |
| Count_NB | 9.94E-01 | 9.73E-01 | 8.19E-01 | 5.35E-01 | 6.86E-06 | 1.33E-04 | 9.09E-06 | 3.58E-01 |
| Count_RF | 7.65E-01 | 9.70E-01 | 9.91E-01 | 9.36E-01 | 2.15E-03 | 2.58E-06 | 6.77E-04 | 4.58E-05 |
| Count_SVM | 1.00E+00 | 9.99E-01 | 9.92E-01 | 9.94E-01 | 7.83E-05 | 6.73E-04 | 1.51E-04 | 7.81E-05 |
| Count_DT | 9.75E-01 | 4.90E-01 | 2.12E-01 | 5.15E-02 | 5.52E-07 | 3.87E-05 | 5.60E-06 | 2.40E-06 |
| Bert_LR | 9.44E-01 | 7.53E-01 | 9.99E-01 | 9.44E-01 | 8.47E-07 | 2.05E-07 | 3.69E-04 | 6.69E-06 |
| Bert_NB | 6.99E-01 | 3.46E-01 | 7.53E-01 | 4.71E-01 | 2.12E-08 | 5.56E-07 | 8.59E-06 | 2.17E-06 |
| Bert_RF | 1.00E+00 | 9.99E-01 | 1.00E+00 | 1.00E+00 | 1.46E-03 | 1.12E-03 | 1.62E-01 | 5.46E-03 |
| Bert_SVM | 1.00E+00 | 9.99E-01 | 1.00E+00 | 1.00E+00 | 2.23E-02 | 5.34E-04 | 2.69E-02 | 2.07E-01 |
| Bert_DT | 9.73E-01 | 8.32E-01 | 4.40E-01 | 3.25E-01 | 0.00E+00 | 3.12E-04 | 1.09E-05 | 6.10E-06 |
| Fast_LR | 6.85E-01 | 5.44E-01 | 9.86E-01 | 7.32E-01 | 9.65E-07 | 2.47E-07 | 1.09E-03 | 1.98E-05 |
| Fast_NB | 9.41E-01 | 6.65E-01 | 9.89E-01 | 8.89E-01 | 1.55E-06 | 2.30E-07 | 6.54E-04 | 3.69E-05 |
| Fast_SVM | 9.92E-01 | 9.92E-01 | 6.57E-01 | 6.09E-01 | 1.83E-05 | 1.69E-04 | 1.91E-05 | 8.41E-06 |
| Fast_DT | 9.70E-01 | 4.87E-01 | 4.86E-01 | 3.18E-01 | 2.19E-05 | 4.83E-06 | 1.14E-05 | 2.21E-06 |
| Bi-Lstm | 8.33E-02 | 7.77E-15 | 1.38E-01 | 3.14E-06 | 1.35E-02 | 2.35E-06 | 3.71E-06 | 1.29E-05 |
| MBERT | 0.00E+00 | 0.00E+00 | 0.00E+00 | 0.00E+00 | 1.99E-07 | 5.99E-07 | 4.02E-05 | 2.02E-06 |

df - degree of freedom

other models. For 3-class English data, t-test results on F1-score show that the improvement is statistically significant. However, the change in features is not significantly improving the performance. For the 5-class English data, random forest with BERT embedding significantly improves the performance compared with all the other models. For 7-class English data, the best model we have chosen is not significantly improving the performance. For all the data sets of Tamil, the best model we have chosen is significantly improving the performance. For 3-class and 7-class Tamil-English, the change in features are not significantly improving the performance, whereas for 5-class Tamil-English, the best model improves the performance significantly.

The classification task is complicated by the relatively small sizer of our dataset and the relatively small proportion of more fine-grained homophobic and transphobic cases. However, these issues are common in other hate speech, offensive language identification and misogynistic cases. In addition, future studies can use the typology and annotation approach we proposed to collect more instances, particularly homophobic and transphobic comments in the non-native script, therefore expanding and balancing the dataset.

**Table 21.** Statistical Test Results for Tamil 7 class dataset

| Models | Chi-Test $\chi^2_{critical}$ (df-4, alpha-0.05) = 0.949 $\chi^2_{statistic}$ of | | | | T - Test | | | |
|---|---|---|---|---|---|---|---|---|
| | Accuracy | Precision | Recall | F1-score | Accuracy | Precision | Recall | F1-score |
| Tf_LR | 9.98E-01 | 8.74E-01 | 9.84E-01 | 9.59E-01 | 2.70E-04 | 4.09E-06 | 4.03E-04 | 7.69E-05 |
| Tf_NB | 9.88E-01 | 8.37E-01 | 1.38E-01 | 5.04E-02 | 5.30E-05 | 1.92E-04 | 1.80E-06 | 6.63E-07 |
| Tf_RF | 8.70E-01 | 9.94E-01 | 9.78E-01 | 9.53E-01 | 1.02E-02 | 1.30E-04 | 5.71E-04 | 2.11E-04 |
| Tf_SVM | 1.00E+00 | 1.00E+00 | 9.83E-01 | 9.89E-01 | 7.28E-03 | 6.75E-04 | 2.61E-04 | 2.34E-04 |
| Tf_DT | 9.79E-01 | 5.22E-02 | 5.08E-02 | 4.33E-03 | 6.48E-05 | 3.56E-06 | 2.42E-06 | 4.53E-07 |
| Count_LR | 9.99E-01 | 9.48E-01 | 9.86E-01 | 9.74E-01 | 9.59E-04 | 1.48E-05 | 3.84E-04 | 1.40E-04 |
| Count_NB | 9.96E-01 | 7.89E-01 | 5.54E-01 | 5.16E-01 | 1.90E-04 | 1.44E-04 | 3.10E-06 | 1.24E-06 |
| Count_RF | 7.55E-01 | 9.83E-01 | 9.63E-01 | 9.26E-01 | 3.35E-03 | 5.18E-04 | 6.19E-04 | 3.71E-04 |
| Count_SVM | 1.00E+00 | 9.99E-01 | 9.80E-01 | 9.87E-01 | 3.96E-03 | 4.19E-03 | 1.94E-04 | 1.99E-04 |
| Count_DT | 9.80E-01 | 4.99E-02 | 5.38E-02 | 4.30E-03 | 4.50E-05 | 1.19E-05 | 2.76E-06 | 8.27E-07 |
| Bert_LR | 9.42E-01 | 5.60E-01 | 1.00E+00 | 9.17E-01 | 1.53E-05 | 8.02E-08 | 1.72E-02 | 5.00E-07 |
| Bert_NB | 4.94E-01 | 5.69E-02 | 5.41E-01 | 1.51E-01 | 1.25E-05 | 2.39E-07 | 1.33E-05 | 3.12E-06 |
| Bert_RF | 1.00E+00 | 9.91E-01 | 1.00E+00 | 1.51E-01 | 9.31E-02 | 5.75E-04 | 1.17E-01 | 3.12E-06 |
| Bert_SVM | 1.00E+00 | 9.98E-01 | 1.00E+00 | 9.99E-01 | 5.00E-01 | 2.24E-03 | 1.83E-02 | 4.09E-03 |
| Bert_DT | 9.83E-01 | 6.63E-01 | 2.51E-01 | 1.65E-01 | 1.86E-04 | 2.61E-05 | 5.69E-06 | 2.02E-06 |
| Fast_LR | 6.01E-01 | 2.34E-01 | 9.82E-01 | 5.09E-01 | 5.58E-06 | 2.69E-07 | 1.86E-04 | 5.73E-06 |
| Fast_NB | 9.45E-01 | 4.75E-01 | 9.83E-01 | 8.10E-01 | 2.34E-05 | 1.06E-08 | 4.12E-05 | 2.22E-06 |
| Fast_SVM | 9.93E-01 | 9.68E-01 | 4.93E-01 | 4.59E-01 | 1.44E-04 | 2.20E-03 | 4.88E-06 | 1.65E-06 |
| Fast_DT | 9.82E-01 | 1.06E-01 | 1.79E-01 | 5.61E-02 | 9.39E-05 | 7.77E-05 | 1.95E-06 | 8.16E-07 |
| Bi-Lstm | 9.93E-02 | 0.00E+00 | 2.76E-02 | 1.68E-08 | 1.59E-02 | 3.93E-07 | 2.15E-06 | 2.45E-07 |
| MBERT | 9.79E-01 | 3.89E-02 | 3.83E-01 | 1.64E-01 | 5.82E-04 | 1.51E-06 | 1.30E-05 | 3.79E-06 |

df - degree of freedom

## 10. Error Analysis

Given the previous sections' results, we can notices that the models' performances in the classification are weak. Thus, this section examines the top models' misclassification errors for each classification task. We give samples of difficult instances (along with their English translation and transliteration). As a result, we look at a total of 27 distinct misclassified comments. A manual analysis allows us to see which category and how misclassification occurs.

The manual analysis of the predictions with the gold labels shows that classification errors are due to four main factors:

- **Weak predictions:** despite the fact that the multilingual comments have been appropriately annotated, the algorithm does not always recognise the homophobia or transphobia category. The model's generalizability may be the cause of these types of errors. It should be noted that the model is only capable of learning a few training instances of some homophobia or transphobia from the fine-grained dataset. It also gets affected by considerable imbalances in the dataset along with code-mixing instances.
- **Mixed labels:** Although few comments are maybe belonging to both homophobia and transphobia, or more than one category in fine-grained settings, the annotators picked the most prevalent one being homophobia. We observe that the model occasionally predicts the other

**Table 22.** Statistical Test Results for Tamil-English 3 class dataset

| Models | Chi-Test $\chi^2_{critical}$ (df-4, alpha-0.05) = 0.949 $\chi^2_{statistic}$ of | | | | T - Test | | | |
|---|---|---|---|---|---|---|---|---|
| | Accuracy | Precision | Recall | F1-score | Accuracy | Precision | Recall | F1-score |
| Tf_LR | 1.00E+00 | 9.76E-01 | 1.00E+00 | 1.00E+00 | 0.00E+00 | 8.55E-02 | 2.68E-02 | 1.45E-02 |
| Tf_NB | 1.00E+00 | 7.77E-01 | 1.00E+00 | 9.99E-01 | 0.00E+00 | 7.78E-04 | 8.37E-03 | 5.07E-03 |
| Tf_RF | 4.88E-03 | 9.63E-01 | 1.00E+00 | 5.04E-01 | 3.53E-02 | 7.32E-02 | 2.28E-01 | 1.78E-02 |
| Tf_SVM | 1.00E+00 | 9.69E-01 | 1.00E+00 | 1.00E+00 | 0.00E+00 | 1.88E-01 | 2.57E-02 | 1.43E-02 |
| Tf_DT | 1.00E+00 | 9.64E-01 | 1.00E+00 | 1.00E+00 | 0.00E+00 | 3.22E-01 | 2.57E-02 | 1.43E-02 |
| Count_LR | 1.00E+00 | 9.92E-01 | 1.00E+00 | 1.00E+00 | 0.00E+00 | 6.17E-02 | 2.30E-02 | 1.23E-02 |
| Count_NB | 1.00E+00 | 7.87E-01 | 1.00E+00 | 9.99E-01 | 0.00E+00 | 3.85E-02 | 1.02E-02 | 6.43E-03 |
| Count_RF | 4.08E-04 | 9.54E-01 | 1.00E+00 | 3.52E-01 | 8.07E-03 | 1.31E-02 | 2.84E-01 | 5.21E-03 |
| Count_SVM | 1.00E+00 | 9.69E-01 | 1.00E+00 | 1.00E+00 | 0.00E+00 | 1.88E-01 | 2.57E-02 | 1.43E-02 |
| Count_DT | 1.00E+00 | 9.58E-01 | 1.00E+00 | 1.00E+00 | 0.00E+00 | 1.14E-01 | 2.24E-02 | 1.22E-02 |
| Bert_LR | 1.00E+00 | 9.94E-01 | 9.76E-01 | 9.95E-01 | 1.29E-06 | 2.08E-02 | 3.92E-05 | 3.84E-04 |
| Bert_NB | 9.15E-01 | 9.63E-01 | 9.78E-01 | 1.00E+00 | 2.28E-06 | 4.66E-03 | 6.04E-04 | 4.43E-01 |
| Bert_SVM | 1.00E+00 | 9.96E-01 | 9.93E-01 | 9.97E-01 | 8.71E-05 | 1.86E-02 | 5.15E-04 | 7.80E-04 |
| Bert_DT | 1.00E+00 | 9.95E-01 | 1.00E+00 | 1.00E+00 | 3.52E-01 | 4.47E-01 | 2.02E-02 | 8.15E-02 |
| Fast_LR | 9.38E-01 | 9.54E-01 | 9.93E-01 | 1.00E+00 | 2.31E-07 | 2.91E-03 | 9.75E-05 | 2.74E-01 |
| Fast_NB | 9.87E-01 | 9.63E-01 | 9.99E-01 | 1.00E+00 | 3.65E-07 | 5.72E-03 | 3.19E-03 | 4.39E-01 |
| Fast_RF | 1.00E+00 | 9.27E-01 | 1.00E+00 | 1.00E+00 | 0.00E+00 | 9.11E-03 | 2.63E-02 | 2.04E-02 |
| Fast_SVM | 1.00E+00 | 9.09E-01 | 1.00E+00 | 1.00E+00 | 0.00E+00 | 1.69E-02 | 2.63E-02 | 1.41E-02 |
| Fast_DT | 1.00E+00 | 8.52E-01 | 1.00E+00 | 9.99E-01 | 0.00E+00 | 7.94E-04 | 4.83E-03 | 3.15E-03 |
| Bi-Lstm | 1.00E+00 | 7.77E-01 | 1.00E+00 | 9.99E-01 | 0.00E+00 | 7.78E-04 | 8.37E-03 | 5.07E-03 |
| BERT | 0.00E+00 | 0.00E+00 | 1.00E+00 | 1.61E-06 | 0.00E+00 | 6.29E-05 | 8.37E-03 | 7.38E-06 |

df - degree of freedom

potential category in these types of mistakes. For instance, the comment: “poda ombothu, neeyelana yethuku irukira” (Translation: “go you nine, why are you living?”) is annotated as “homophobia-threatening” because the authors are offending the vulnerable target individual by asking them why are you living- in other sense provoking them to commit suicide. The system predicted as ”transphobic-threatening”, this is also true since ‘nine’ is used as a derogatory word to insult all the LGBT+ people. However, the annotators choose homophobia based on the previous comments. Since this information was not available to simple classifiers, our results were low.

- **Unique phrases:** Some comments curse and insult LGBT+ vulnerable individuals in a unique style or with a word. The Tamil language, like other languages, has several expressions for cursing or insulting. Aside from well-known words, some authors coined their own expressions that require cultural understanding to comprehend in both Tamil native script and code-mixed settings. An example of these cases is the comment: “ivan dosthana padathil varuvaran poola irkirane”. Literal translation, he looks like a person from the Dostana movie. This comment is not a homophobic sentence; however, the main characters pretend to be gay to get a house in the Dostana movie. Therefore, this comment’s indirect meaning is that the author is insulting the other person by saying you are looking gay.

**Table 23.** Statistical Test Results for Tamil-English 5 class dataset

| Models | Chi-Test $\chi^2_{critical}$ (df-4, alpha-0.05) = 0.949 $\chi^2_{statistic}$ of | | | | T - Test | | | |
|---|---|---|---|---|---|---|---|---|
| | Accuracy | Precision | Recall | F1-score | Accuracy | Precision | Recall | F1-score |
| Tf_LR | 1.00E+00 | 9.99E-01 | 1.00E+00 | 9.50E-01 | 8.07E-03 | 1.06E-01 | 7.58E-03 | 2.06E-02 |
| Tf_NB | 1.00E+00 | 1.99E-01 | 9.97E-01 | 9.70E-01 | 1.94E-03 | 1.98E-05 | 4.16E-04 | 1.90E-04 |
| Tf_RF | 1.04E-02 | 9.90E-01 | 1.00E+00 | 8.29E-01 | 3.41E-02 | 1.50E-01 | 7.15E-03 | 1.31E-02 |
| Tf_SVM | 1.00E+00 | 9.99E-01 | 1.00E+00 | 9.98E-01 | 1.87E-01 | 2.07E-01 | 6.05E-03 | 3.41E-03 |
| Tf_DT | 1.00E+00 | 9.31E-01 | 9.99E-01 | 9.93E-01 | 8.89E-02 | 2.97E-02 | 2.58E-03 | 1.96E-03 |
| Count_LR | 1.00E+00 | 9.99E-01 | 1.00E+00 | 9.99E-01 | 8.07E-03 | 2.18E-02 | 7.58E-03 | 3.44E-03 |
| Count_NB | 1.00E+00 | 5.93E-01 | 9.97E-01 | 9.78E-01 | 1.94E-03 | 3.53E-02 | 4.16E-04 | 2.96E-04 |
| Count_RF | 6.45E-02 | 9.99E-01 | 1.00E+00 | 9.23E-01 | 8.49E-02 | 2.45E-02 | 7.15E-03 | 1.23E-02 |
| Count_SVM | 1.00E+00 | 9.98E-01 | 1.00E+00 | 9.97E-01 | 8.89E-02 | 2.36E-01 | 6.05E-03 | 3.01E-03 |
| Count_DT | 1.00E+00 | 9.08E-01 | 9.99E-01 | 9.92E-01 | 8.89E-02 | 8.81E-04 | 2.58E-03 | 4.17E-04 |
| Bert_LR | 9.98E-01 | 9.66E-01 | 9.71E-01 | 9.97E-01 | 9.34E-06 | 6.26E-04 | 2.54E-05 | 1.49E-03 |
| Bert_NB | 7.99E-01 | 8.51E-01 | 9.81E-01 | 1.00E+00 | 2.82E-06 | 2.22E-04 | 3.73E-04 | 1.51E-01 |
| Bert_SVM | 1.00E+00 | 9.79E-01 | 9.94E-01 | 9.98E-01 | 1.16E-03 | 9.43E-04 | 3.42E-05 | 5.56E-04 |
| Bert_DT | 1.00E+00 | 9.80E-01 | 1.00E+00 | 1.00E+00 | 2.32E-03 | 6.25E-03 | 6.49E-02 | 3.42E-02 |
| Fast_LR | 8.32E-01 | 8.38E-01 | 9.86E-01 | 1.00E+00 | 7.14E-08 | 9.60E-05 | 4.64E-06 | 2.09E-01 |
| Fast_NB | 9.89E-01 | 8.53E-01 | 9.98E-01 | 1.00E+00 | 5.41E-07 | 1.18E-04 | 2.47E-05 | 3.63E-01 |
| Fast_RF | 1.00E+00 | 9.12E-01 | 1.00E+00 | 9.98E-01 | 0.00E+00 | 8.54E-03 | 2.12E-03 | 1.09E-01 |
| Fast_SVM | 1.00E+00 | 9.37E-01 | 9.99E-01 | 9.93E-01 | 4.97E-02 | 3.35E-02 | 1.14E-03 | 7.35E-02 |
| Fast_DT | 1.00E+00 | 2.01E-01 | 9.97E-01 | 9.45E-01 | 1.94E-03 | 2.04E-05 | 3.60E-04 | 1.27E-04 |
| Bi-Lstm | 1.54E-11 | 0.00E+00 | 9.97E-01 | 8.06E-05 | 1.50E-10 | 6.15E-06 | 4.16E-04 | 8.91E-06 |
| BERT | 5.30E-14 | 2.75E-04 | 1.00E+00 | 3.71E-01 | 1.78E-09 | 6.53E-06 | 5.29E-02 | 7.84E-05 |

df - degree of freedom

- **False homophobic or transphobic cases:** Our dataset contains both anti-LGBT+ and non-anti-LGBT+ comments to categorize both groups, since we are interested in detecting anti-LGBT+ comments from social media. Many non-anti-LGBT+ are neutral (e.g., thoughts, inquiries, news), but some general insulting comments aren't directed towards LGBT+. Systems sometimes predict these comments as one of the anti-LGBT+ comments.

Table 10 shows the examples of misclassified comments in English language settings. From the Table, we can see that some of the non-anti-LGBT+ contents are predicted as transphobic or homophobic due to the simple mention of Lesbian or gay. We have also shown the examples of 'Hope Speech' and 'Counter speech' wrongly predicted as 'None-of-the-above' in the 5-class problem. This kind of error makes any mention of LGBT+ in comments as anti-LGBT+, and if the system were to be automated, this marginalises the LGBT+. In the future, people who use the dataset should consider these errors so that the LGBT+ supportive comments are not wrongly penalised.

Table 10 shows the examples of misclassified comments in Tamil-English code mixed settings with English translation. We have shown that some of the threatening comments are classified as none-of-the-above. One such example of asking LGBT+ vulnerable individuals to take life is misclassified as none-of-the-above. As this is a serious issue, the future models developed for

**Table 24.** Statistical Test Results for Tamil-English 7 class dataset

| Models | Chi-Test $\chi^2_{critical}$ (df-4, alpha-0.05) = 0.949 $\chi^2_{statistic}$ of | | | | T - Test | | | |
|---|---|---|---|---|---|---|---|---|
| | Accuracy | Precision | Recall | F1-score | Accuracy | Precision | Recall | F1-score |
| Tf_LR | 1.00E+00 | 9.76E-01 | 1.00E+00 | 1.00E+00 | 4.97E-02 | 1.76E-02 | 1.03E-03 | 7.93E-04 |
| Tf_NB | 1.00E+00 | 2.75E-01 | 9.99E-01 | 9.89E-01 | 1.71E-02 | 2.62E-03 | 3.32E-04 | 2.78E-04 |
| Tf_RF | 7.65E-06 | 9.95E-01 | 1.00E+00 | 8.09E-01 | 2.29E-09 | 3.62E-02 | 8.76E-03 | 6.60E-06 |
| Tf_SVM | 1.00E+00 | 9.95E-01 | 1.00E+00 | 9.99E-01 | 5.00E-01 | 1.39E-01 | 1.02E-03 | 3.77E-03 |
| Tf_DT | 1.00E+00 | 9.28E-01 | 1.00E+00 | 9.98E-01 | 3.11E-01 | 8.83E-03 | 3.92E-04 | 3.93E-04 |
| Count_LR | 1.00E+00 | 9.56E-01 | 1.00E+00 | 1.00E+00 | 3.52E-02 | 2.42E-02 | 5.29E-04 | 3.92E-04 |
| Count_NB | 1.00E+00 | 5.82E-01 | 9.99E-01 | 9.92E-01 | 1.71E-02 | 7.06E-02 | 6.42E-04 | 6.26E-04 |
| Count_RF | 5.07E-06 | 9.95E-01 | 1.00E+00 | 8.42E-01 | 2.95E-09 | 1.22E-01 | 6.93E-03 | 5.58E-06 |
| Count_SVM | 1.00E+00 | 9.95E-01 | 1.00E+00 | 9.99E-01 | 1.87E-01 | 1.31E-01 | 1.15E-03 | 1.50E-03 |
| Count_DT | 1.00E+00 | 9.28E-01 | 1.00E+00 | 9.98E-01 | 3.11E-01 | 8.83E-03 | 3.92E-04 | 3.93E-04 |
| Bert_LR | 9.97E-01 | 9.72E-01 | 9.64E-01 | 9.93E-01 | 4.43E-06 | 5.02E-02 | 3.36E-05 | 4.11E-04 |
| Bert_NB | 7.54E-01 | 8.76E-01 | 9.77E-01 | 1.00E+00 | 1.64E-06 | 1.50E-02 | 3.23E-04 | 2.86E-01 |
| Bert_SVM | 1.00E+00 | 9.80E-01 | 9.91E-01 | 9.95E-01 | 1.03E-03 | 1.02E-01 | 2.96E-04 | 1.89E-03 |
| Bert_DT | 1.00E+00 | 9.52E-01 | 1.00E+00 | 1.00E+00 | 8.89E-02 | 3.95E-02 | 2.37E-01 | 3.02E-01 |
| Fast_LR | 7.51E-01 | 8.77E-01 | 9.76E-01 | 1.00E+00 | 8.41E-08 | 9.96E-03 | 1.24E-06 | 1.22E-01 |
| Fast_NB | 9.89E-01 | 9.02E-01 | 9.98E-01 | 1.00E+00 | 5.41E-07 | 1.02E-02 | 9.21E-06 | 4.11E-02 |
| Fast_RF | 1.00E+00 | 9.33E-01 | 1.00E+00 | 9.99E-01 | 1.71E-02 | 3.72E-02 | 2.92E-03 | 3.29E-03 |
| Fast_SVM | 1.00E+00 | 9.68E-01 | 1.00E+00 | 9.99E-01 | 8.89E-02 | 1.07E-01 | 3.34E-02 | 2.29E-02 |
| Fast_DT | 1.00E+00 | 3.32E-01 | 9.99E-01 | 9.90E-01 | 1.71E-02 | 2.24E-03 | 2.64E-04 | 2.28E-04 |
| Bi-Lstm | 1.00E+00 | 2.75E-01 | 9.99E-01 | 9.89E-01 | 1.71E-02 | 2.62E-03 | 3.32E-04 | 2.78E-04 |
| BERT | 0.00E+00 | 6.09E-03 | 9.95E-01 | 3.29E-01 | 2.51E-08 | 1.54E-03 | 5.93E-02 | 2.10E-04 |

df - degree of freedom

this dataset should be more sensitive to these issues. As there is no standard spelling for the code-mixing content, some important cases are misclassified.

Table 10, shows the examples of misclassified comments in Tamil language settings with English translation. We have shown the examples of homophobic and transphobic comments are predicted as “Counter speech” or “None-of-the-above”; these are also influenced by cultural knowledge. For future works, then we should also try to bring cultural knowledge into the system so that our system predictions are more accurate. Some of the sentences are very subtle, which does not seem homophobic/transphobic but they are homophobic/transphobic by implying heterosexual is the normal and acceptable one.

## 11. Conclusion

We present a dataset with the high-quality, expert classification of homophobic and transphobic content from multilingual YouTube comments, as well as a hierarchical granular taxonomy for homophobia and transphobia in this study. The resulting dataset is negligible in contrast to other annotated datasets used for other classification. However, to the best of our knowledge, this is the first dataset to be created for homophobia and transphobia in multilingual comments in Tamil, English and Tamil-English. We conducted a comprehensive empirical study by assessing a range of feature selection approaches, including machine and deep learning methods, inside a supervised classification framework. We trained a large number of classifiers on 3-class, 5-class, and 7-class datasets with diverse feature spaces. We also did manual corpora analysis and error analysis to understand the nature of the task. Our work demonstrated that homophobic and transphobic detection is a challenging problem to be solved in multilingual and multicultural settings.

**Table 25.** Error Analysis for English

| | Original Label | Comment | Prediction |
|---|---|---|---|
| English 3 class | Non-anti-LGBT+ content | Lesbian sex is important in our society | Transphobic |
| English 3 class | Non-anti-LGBT+ content | Many may know such areas and such problems like this bt they wont care bt for my opinion i have given a point thats all | Homophobic |
| English 3 class | Homophobic | Mental speaking and mentals listening this is future world ending | Transphobic |
| English 5 class | Counter-speech | Correct she is good heart lady not only she all transgender is really good heart person public should understand them | None-of-the-above |
| English 5 class | Hope-Speech | Hamelton Melvin superif you need any help to your service contact meiif I can do it I will help to you | None-of-the-above |
| English 5 class | None-of-the-above | I like your videos. They are great. | Hope-Speech |
| English 7 class | None-of-the-above | Why do you put the title as transgender For views | Counter-speech |
| English 7 class | None-of-the-above | We are waiting for part 2 | Hope-speech |
| English 7 class | Hope-Speech | Nice suggestion bro needed for whole India itself | Counter-speech |

Our work has several future directions; we are interested in creating the dataset for other Dravidian languages. Also, we would like to considerably increase the size of the dataset for Tamil by crawling and annotating more data set from social media. We will also plan to explore both semi-supervised approaches and incremental approaches to improve the performance of the classifiers. Moreover, since we found many anti-LGBT+ comments are sarcastic in our manual analysis, we plan to study the correlation between sarcasm and anti-LGBT+ comments.

## Competing interests:

The author(s) declare none

**Table 26.** Error Analysis for Tamil-English

| | Original Label | Comment | Prediction |
|---|---|---|---|
| Tamil-English 3 class | Homophobic | Enna krumam da Idhu Translation: What the shit is that? | Transphobic |
| Tamil-English 3 class | Transphobic | Saniyanungala savadikanum ithungala intha Uzhagathula irukanum nu thevaiya illa Translation: Bad omans we should kill them all they dont deserve to be in the world | Counter-speech |
| Tamil-English 3 class | Non-anti-LGBT+ content | Pooo dd loosuuu over pesathaa but CRT pesuu try pannuuu Translation: you are crazy dont talk too much but try to speak correctly | Homophobic |
| Tamil-English 5 class | None-of-the-above | Kattupaadu suthanthiram na enna broo Neenga sollunglaaen keppom Translation: Tell me what is freedom and what is protection | Hope-Speech |
| Tamil-English 5 class | Homophobic | Epdiyo poi tholainga Ipo unak enna vaenum sex pananum panika Translation: whatever you go and do, you want to have sex now fag, then go have sex | Counter-speech |
| Tamil-English 5 class | None-of-the-above | Yedukurathu pitcha ithula thimuru vera ivanguluku Translation: they are begging but they are harassing | Homophobic |
| Tamil-English 7 class | Homophobic-Threatening | Chi karumam Ena di nyayam pesura naadu naasama poirum Di Culture poirum ena pechu pesura ne unga Amma vum ponnu dhan marandhutu pesadha chi saavu di poitu Translation: What are you talking everything will go bad, your mother and father are in pain you being fag, you should kill yoursefl | None-of-the-above |
| Tamil-English 7 class | None-of-the-above | Antha thirunangai ku avan vali purigirathu athanalaa ivlo convincing panraanga Translation: That transgender understand his pain, she is helping him so much | Counter-speech |
| Tamil-English 7 class | Transphobic-Threatening | Entha Mari aravaningala seruppala adikkanum entha Mari prachanai Ella edathulayum erukku pothumakkalukku pathukappu illa Translation: We should beat these kind of transgenders with boot, all the problems are in trans community as well | None-of-the-above |

**Table 27.** Error Analysis for Tamil

| | Original Label | Comment | Prediction |
|---|---|---|---|
| Tamil 3 class | Non-anti-LGBT+ content | நீங்க தான் சம்பளம் ஒருநாை◌ளக்கு 1500ருபாய் ே◌கக்குறிங்க Translation: You ask 1500 rupees per day salary | Transphobic |
| Tamil 3 class | Transphobic | ே◌பாடி ெ◌மண்டல் ஒன்பது Translation: You mental nine get lost | Counter-speech |
| Tamil 3 class | Transphobic | பஸ்ல இவங்க ெ◌தால்ை◌ல தாங்க முடியைல 10 ருபாய் ெ◌காடுத்ே◌த ஆகணும் இல்ை◌லனா அசிங்கமா திட்டி சாபம் விடுறாங்க Translation: They are nuisance in bus, they ask for 10rupees, if we dont give them then curse us badly | Non-anti-LGBT+ content |
| Tamil 5 class | None-of-the-above | இந்த மாதிரி திருநங்ை◌க நால எல்லாத்துக்கும் ெ◌கட்ட ே◌பரு Translation: Because of the these bad transgender, whole transgendercommunity get bad name | Counter-speech |
| Tamil 5 class | Hope-Speech | நல்லேத நடக்கும் சேகாதரி வாழ்த்துெ◌மன்ே◌மலும் வளர Translation: Everything will be okay sister, you will succeed in future | None-of-the-above |
| Tamil 5 class | Homophobic | மக்கைள பிரிக்கின்ற ெ◌சயல் நான் ஏற்க மாட்ே◌டன் இனம் மதம் நாடு ே◌பான்று இதுவும் பிரிவிை◌ன தான் நாங்கள் ைஅனவரும் நம் அம்மா அப்பா மூலம் தான் வந்திருக்கிே◌றாம் இந்த உலகிற்கு அவர்களும் ஓரினர் ஆக இருந்திருந்தால் நாங்கள் பிறந்திருக்க மாட்ே◌டாம்ைஆச அழிை◌வத் தான் தரும் Translation: I do not accept the act of dividing people like race and religion. This is also division. We have all come through our mother and father. | None-of-the-above |
| Tamil 7 class | Transphobic-derogation | பண்ணி தன்ே◌னாட குணத்ை◌த எப்பவுே◌ம மாற்றாது இது ஒரு ஈனப்பிறவி ெ◌ஜன்மங்கள் Translation: Pig never changes its character, it is a genetic genes | None-of-the-above |
| Tamil 7 class | Homophobic-derogation | முடிவு காலம் இப்படித்தான் நடக்கும் Translation: This is how the end times will be | Counter-speech |
| Tamil 7 class | None-of-the-above | ஒருத்தனாவது எதிர்த்து ே◌கள்விே◌கக்குறானா பாத்தீங்களா Translation: Did not see, no one has questions | Hope-Speech |